\pdfoutput=1

\documentclass[11pt]{article}
\usepackage{multirow}
\usepackage{graphicx}
\usepackage{pdflscape}
\usepackage{float}
\usepackage{threeparttable}
\usepackage{stfloats}
\usepackage{tablefootnote}
\usepackage[table,xcdraw]{xcolor}
\usepackage{acl}

\usepackage{times}
\usepackage{latexsym}

\usepackage[T1]{fontenc}

\usepackage[utf8]{inputenc}

\usepackage{microtype}

\usepackage{inconsolata}

\usepackage{graphicx}
\usepackage[normalem]{ulem}
\useunder{\uline}{\ul}{}
\usepackage{ulem}
\usepackage{booktabs}  
\usepackage{footnote} 

\title{Oedipus and the Sphinx: Benchmarking and Improving Visual Language Models for Complex Graphic Reasoning}

\author{
  Jianyi Zhang$^*$, Xu Ji, Ziyin Zhou, Yuchen Zhou, Shubo Shi, Haoyu Wu, Zhen Li, Shizhao Liu \\
  \\  
  Beijing Electronic Science \& Technology Institute \\
  \\  
  $^*$Corresponding author: \texttt{zjy@besti.edu.cn}
}

\begin{document}
\maketitle
\begin{abstract}

Evaluating the performance of visual language models (VLMs) in graphic reasoning tasks has become an important research topic. However, VLMs still show obvious deficiencies in simulating human-level graphic reasoning capabilities, especially in complex graphic reasoning and abstract problem solving, which are less studied and existing studies only focus on simple graphics. To evaluate the performance of VLMs in complex graphic reasoning, we propose \textbf{ReasonBench}, the first evaluation benchmark focused on structured graphic reasoning tasks, which includes 1,613 questions from real-world intelligence tests. ReasonBench covers reasoning dimensions related to location, attribute, quantity, and multi-element tasks, providing a comprehensive evaluation of the performance of VLMs in spatial, relational, and abstract reasoning capabilities. We benchmark 11 mainstream VLMs (including closed-source and open-source models) and reveal significant limitations of current models. Based on these findings, we propose a dual optimization strategy: \textbf{Diagrammatic Reasoning Chain (DiaCoT)} enhances the interpretability of reasoning by decomposing layers, and \textbf{ReasonTune} enhances the task adaptability of model reasoning through training, all of which improves VLM performance by 33.5\%. All experimental data and code are in the repository: \url{https://huggingface.co/datasets/cistine/ReasonBench}.

\end{abstract}

\section{Introduction}

\begin{figure}
    \centering
    \includegraphics[width=1\linewidth]{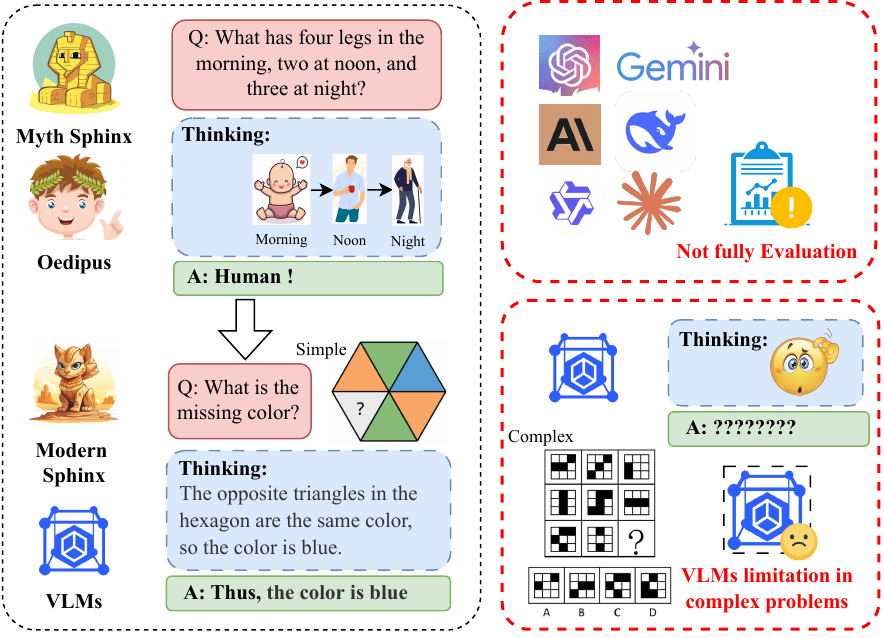}
    \caption{The Oedipus and the Sphinx in mythology can be analogized to VLMs handling of graphical reasoning problems.  However, VLMs appear to struggle with complex graphical reasoning and lack comprehensive evaluation.}
    \label{fig:Sphnix}
\end{figure}

The rapid development of visual language models (VLMs) is reshaping the field of AI, enabling them to jointly understand and reason across visual and textual modalities for the first time. VLMs have made significant advancements on tasks such as image captioning\cite{Vision-and-language_navigation_CVPR2018} and open-domain visual question answering\cite{VQA_Antol_ICCV2015}. However, current benchmark research remains fragmented. Existing datasets (e.g., Raven\cite{RAVEN_dataset_2019CVPR} and CLEVR\cite{CLEVER_datasett_2017CVPR}) lack diversity and cannot fully evaluate reasoning capabilities. Recent studies have explored dynamic and abstract reasoning, but fail to capture the true complexity of graphical reasoning. We can't help but wonder, when the Sphinx's riddle shifts to complex graphic reasoning (Figure \ref{fig:Sphnix}), can current VLM play the role of Oedipus and solve the Sphinx's riddle? Simple graphical reasoning refers to the case where there are only one or a few elements in the graph\cite{PuzzleVQA_ACL2024}, or the model is used to solve simple geometric angle problems\cite{deng2024r}. Complex graphical reasoning refers to the case where there are dimensions such as position, style, attribute, ablation, etc. between graphs, including intersections between these dimensions. And human will be elusive and cramped when dealing with such reasoning. To investigate VLMs' performance in complex graphical reasoning, we perform a comprehensive analysis of existing benchmarks. Our findings indicate that their capability to tackle complex graphical reasoning remains underexplored, as previous studies have primarily concentrated on simple graphical reasoning. To fill this research gap, we systematically study the complex graphical reasoning capabilities of VLMs in structured settings for the first time, and introduce ReasonBench, a diagnostic benchmark specifically designed to evaluate structured graphical reasoning.

ReasonBench consists of 1,613 standardized test problems and integrates 11 core cognitive reasoning dimensions, including Positional, Stylistic, Attribute, Quantitative, Spatial, and other related dimensions. We systematically evaluate 11 VLMs, which include both closed-source and open-source ones. Additionally, we introduce a triple-controlled evaluation protocol to ensure measurement reliability and cross-model comparability. We also establish a human-level performance baseline, where human participants (mean age = 28.3 ± 6.5 years) achieve an average accuracy of 68.7\% on the same test set, offering a crucial reference for future model improvements.

Based on these findings, we identified several key trends that underscore the primary limitations of current VLMs: (1) Even the best-performing models achieve only an average accuracy of 27\%. (2) The performance gap between integrated and separated format for VLMs is minimal. Overall, the gap between closed-source and open-source models has not significantly widened.

To enhance VLMs’ graphical reasoning capabilities. First, we introduce \textbf{Diagrammatic Chain-of-Thought (DiaCoT)}, which leverages visualized, step-by-step reasoning from a layered perspective of the graph to improve interpretability. Second, we propose a fine-tuning strategy \textbf{ReasonTune}, which gradually strengthens models’ inductive reasoning abilities. Experimental results demonstrate that our optimization framework improves complex graphical reasoning accuracy by 33.5\%. Hence, our main contributions include:
\begin{enumerate}
    \item We present ReasonBench, the first benchmark designed specifically for evaluating complex graphical reasoning abilities of VLMs, covering 11 cognitive dimensions and 29 task types. The dataset addresses the lack of real-world applicability of previous benchmarks using our designed protocol. In addition, we provide a human performance baseline.
    
    \item We systematically exam 11 closed-source and open-source VLMs, first revealing key limitations in current VLMs and elaborating a series of intriguing findings that highlight the importance of structured evaluation for graphical reasoning.
    
    \item We propose a dual optimization framework to enhance VLMs' graphical reasoning abilities, accommodating both open-source and closed-source models. Our DiaCoT and ReasonTune strategies significantly improve performance, achieving a 33.5\% accuracy boost in complex graphical reasoning tasks.
\end{enumerate}

\begin{figure*}[!t]
    \centering
    \includegraphics[width=\linewidth]{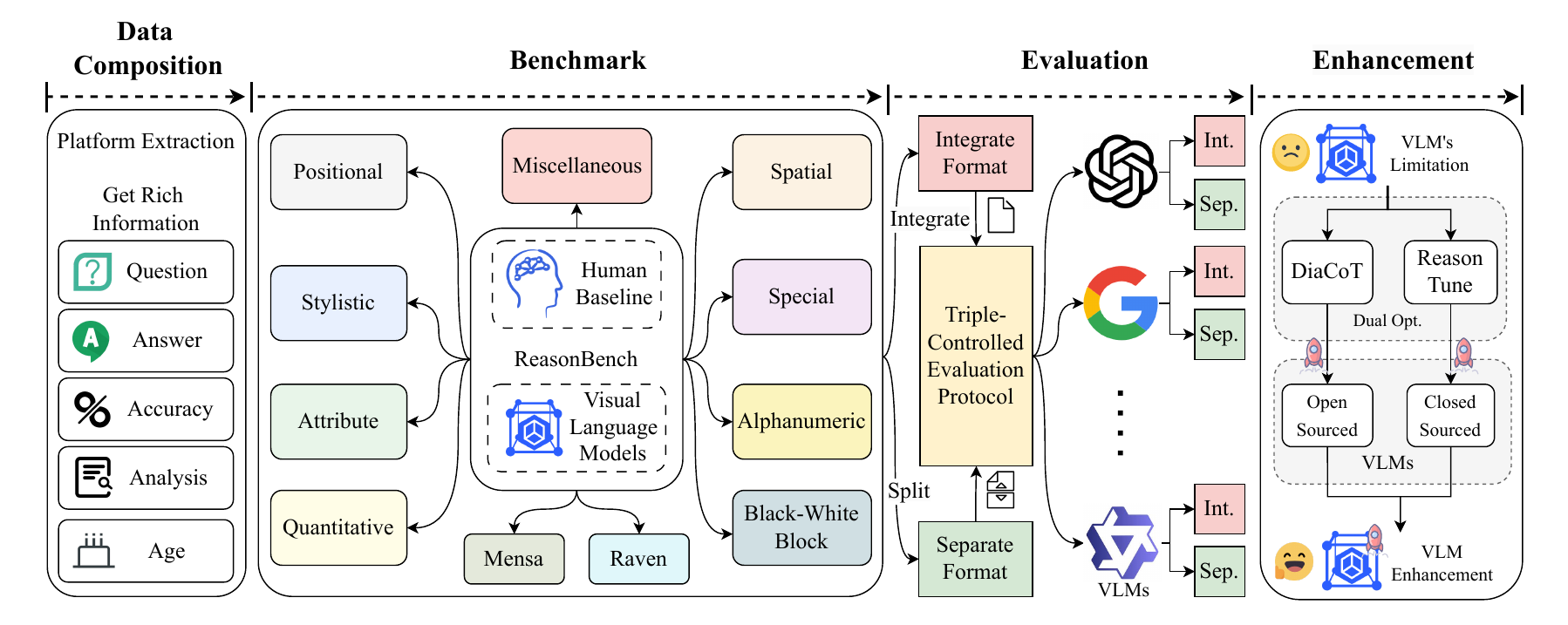}
    \caption{The pipeline of our work. First, in the \textbf{data combination} phase, we obtain rich information. Second, we propose the first \textbf{benchmark} ReasonBench from 11 cognitive dimensions. Third, in the \textbf{evaluation} phase, we evaluate the reasoning ability of VLM by integrating and separating formats. Experiments show that our approach can enhance the ability of VLM.}
    \label{fig:pipeline}
\end{figure*}

\section{Related Work}
Recent advances in VLMs have enabled significant progress in multimodal tasks such as image captioning\cite{Vision-and-language_navigation_CVPR2018} and visual question answering\cite{VQA_Antol_ICCV2015}. However, evaluating VLMs’ capabilities in structured, complex graphical reasoning remains underexplored. Early benchmarks like Raven\cite{RAVEN_dataset_2019CVPR}, focused on matrix completion tasks, but lacked diversity in reasoning types. CLEVR\cite{CLEVER_datasett_2017CVPR} emphasized compositional logical reasoning but diverged from real-world testing scenarios. Recent efforts, such as Mementos\cite{Mementos_ACL2024} for sequential dynamic reasoning and PuzzleVQA\cite{PuzzleVQA_ACL2024}is used to solve single or double elements  reasoning, expanded the scope but still fell short of capturing the complexity of graphical reasoning problems.

Human intelligence tests, such as Mensa and RPM exams, have long served as gold standards for evaluating abstract reasoning. Prior work adapted RPM-style tasks to assess machine intelligence\cite{zhang2019learning}, but these efforts often lacked real-world applicability and fine-grained cognitive diagnostics. The Raven dataset\cite{RAVEN_dataset_2019CVPR} primarily focuses on RPM-style matrix completion but lacks diversity in problem types, thus constrain its ability to evaluate broader reasoning capabilities. The CLEVR dataset\cite{CLEVER_datasett_2017CVPR} emphasizes compositional logical reasoning but deviates significantly from real-world testing scenarios.  Existing evaluations\cite{Beyond_accuracy_ACL2020} prioritized accuracy over limiting insights into models' failure modes. Recent studies also identified weaknesses in VLMs’ spatial and relational reasoning, particularly in tasks included occlusion, composite transformations, and 3D geometric operations\cite{camel_NIPS2023}.  Although existing benchmarks and methods have advanced the field, they exhibit fragmented task coverage, limited real-world relevance, and insufficient diagnostic granularity. For instance, video-centric benchmarks like ActivityNet-Captions\cite{krishna} and Charades\cite{sigurdsson} focused on action recognition rather than structured reasoning. Similarly, HellaSwag\cite{hellaswag}, though primarily text-based, inspired multimodal extensions but lacked alignment with complex graphical reasoning problem-solving. Therefore, our work bridges these gaps by introducing ReasonBench, the first benchmark derived from standardized human intelligence tests.

To address reasoning limitations, methods like Chain-of-Thought (CoT) prompting\cite{COT} and neuro-symbolic approaches\cite{liang2024can} have been proposed. CoT improved the interpretability of the model in text-based reasoning, but struggled with visual complexity. Neuro-symbolic frameworks integrated symbolic rules with neural networks, but required extensive domain-specific engineering. Fine-tuning strategies, such as progressive training on synthetic data\cite{CLEVER_datasett_2017CVPR}, enhanced specific capabilities but failed to generalize across diverse reasoning dimensions. Therefore, we propose DiaCoT and ReasonTune, a dual optimization framework that combines visualized reasoning chains with structured fine-tuning to address the cognitive bottlenecks of VLMs. while addressing their limitations in multimodal, complex graphical reasoning reasoning scenarios.

\section{ReasonBench}

\begin{table*}[!t]
  \vspace*{-10pt} 
  \centering
  \includegraphics[width=\textwidth]{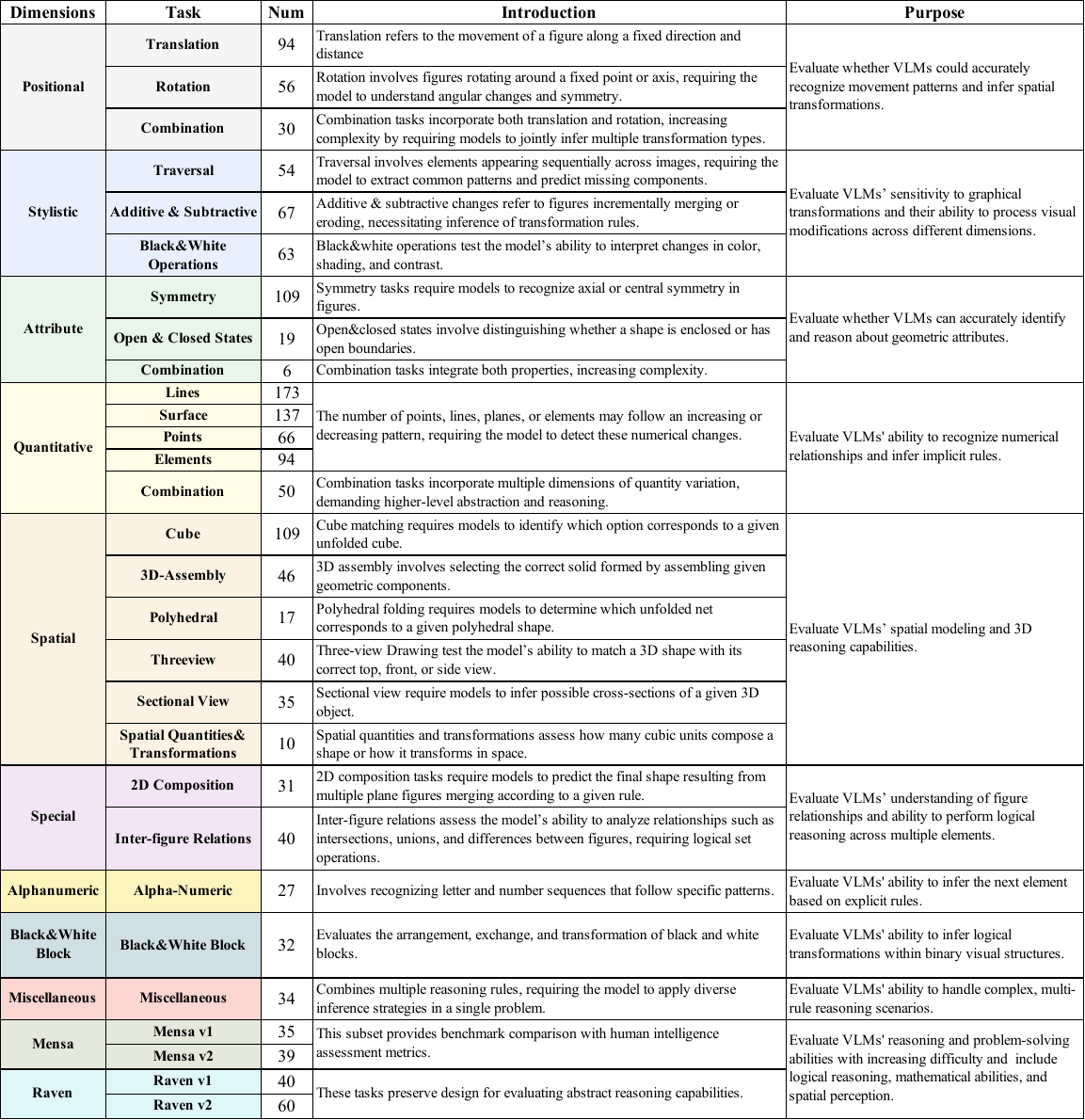}
  \caption{The overview of ReasonBench. Different colors represent different \textbf{Dimensions}. \textbf{Task} indicates one or more tasks under each dimension. \textbf{Num} refers to the number of test questions for each task. \textbf{Introduction} provides a brief overview of each task. \textbf{Purpose} describes the utility and impact of the different dimensions we designed.}
  \label{fig:dataset}
\end{table*}

\subsection{Dataset Composition}
ReasonBench is a meticulously curated dataset consisting of 1,613 real-world graphical reasoning problems, systematically gathered from three primary sources: Chinese Civil Service Aptitude Tests\cite{CMMLU}, Mensa Intelligence Tests\footnote{\url{https://www.mensa.org/}}, and Raven's Progressive Matrices\footnote{\url{https://psycho-tests.com/test/raven-matrixes-test}}. The dataset includes 11 cognitive dimensions, which are further categorized into 29 distinct task types. This multi-source hierarchical architecture enables comprehensive evaluation of VLMs analytical capabilities across diverse reasoning paradigms. Table \ref{fig:dataset} shows the details.

\subsection{Human Performance}
To further explore the comparison between VLMs and human reasoning abilities, we extracted data from existing resources. We collect accuracy results for 1439 questions from the FenBi\footnote{\url{https://fenbi.com/}} platform, with participants in the age group 21 to 35, including the questions themselves, correct answers, human accuracy, and detailed analysis. Additionally, for the Mensa and Raven tests, we invited three participants aged from 22 to 25, who voluntarily completed the tests with the consent of their respective mentors. Our analysis revealed an average baseline human score of 69.76\%.

\subsection{Evaluation Procedure and Metrics}
Our experimental framework systematically evaluates the reasoning capabilities of VLMs on complex graphical reasoning tasks. As illustrated in Figure \ref{fig:pipeline} Evaluation phase. We tested 11 VLMs in batches on ReasonBench. Meanwhile, we investigated the different effects of integrated and segmented visual presentation formats on VLM. Due to space limitations, the specific format is shown in Appendix \ref{ref:Int&Sep}).
\begin{itemize}
    \item \textbf{Integrated format}: The problem and answer choices are presented in a single graphic.
    \item \textbf{Separated format}: The problem and answer choices are splited into multiple graphics and sequentially fed into the model.
\end{itemize}
We apply the Triple-controlled Evaluation Protocol (see Section \ref{tex:triple}) to both formats to address the lack of real-world applicability of previous benchmarks. Finally, we use automatic keyword extraction to systematically analyze model responses. This evaluation is used to measure the reasoning ability of different VLMs on graphics using the accuracy (ACC) metric. For examples of each task, see Appendix \ref{example}.

\subsection{Triple-controlled Evaluation Protocol}
\label{tex:triple}
In the task design phase of our evaluation, we systematically explored various response formats. Through an in-depth analysis of the characteristics of visual reasoning tasks, we identify significant modality differences between graphic-presented questions and options and the text-based responses generated by the model. To mitigate the impact of non-cognitive factors on the validity of the evaluation, we established a standardized evaluation protocol with triple control, as outlined below:

\textbf{1.} We use multiple choice question (MCQ) as the evaluation question type. Except for the Mensa IQ test and Raven's reasoning matrix, which retain 6-8 options, all other questions are designed with four options. At the same time, we ensure that the frequency fluctuation of each correct option (A/B/C/D) in all questions is controlled within 25\% ± 0.7\% (see Appendix \ref{MCQ_distribution}).

\textbf{2.} Both integrated and separated formats use distinct, fixed templates. A fixed prompt template is used to ensure consistent input formatting for each question, while a structured response format (\textbf{\texttt{<ans>OptionX</ans>}}) is applied to minimize the impact of non-cognitive factors on model outputs (see Appendix \ref{format}). Regardless of the format type, whether integrated or separated, graphics are converted into a universally accessible API URL format to ensure compatibility across all models, with all data being made publicly available.

\textbf{3.} A standardized evaluation protocol based on Pass@1 single-attempt scoring \cite{pass1_ICLR2021} is applied to ensure the objectivity and consistency of the evaluation results (see Appendix \ref{pass@k}).

\section{Experiments}
In our experimental analysis, we explore two key Research Questions:

\textbf{RQ1: How VLMs perform in complex graphical reasoning?} We evaluate closed-source and open-source VLMs on complex graphic-based problems, assessing their reasoning capabilities across multiple cognitive dimensions.

\textbf{RQ2: What is the impact of input strategies?} Using a separated format, we sequentially present independent graphical dimensions to analyze input format effects on model performance.

\subsection{Models}
This study is based on Chatbot Arena Leaderboard\footnote{\url{https://huggingface.co/spaces/lmarena-ai/chatbot-arena-leaderboard}}, an authoritative evaluation ranking published on HuggingFace. A total of 11 representative VLMs were selected to construct the benchmarking framework. The selected models are categorized into closed-source and open-source groups, strictly adopting the best-performing versions officially validated within each model series as of December 25, 2024.
The closed-source group includes six commercial systems: GPT-4o\cite{GPT-4o}, Gemini-2.0\cite{Gemini}, Gemini-1.5\cite{Gemini-1.5}, Claude-3.5\cite{claude3}, GLM-4V\cite{GLM-4V}, and Yi-Vision\cite{yi}. The open-source group consists of five reproducible implementations: Qwen-VL-72B\cite{Qwen-VL}, InternVL2-26B\cite{Internvl}, Pixtral\cite{Pixtral}, QVQ-72B-Preview\cite{Qwen2-VL}, and DeepSeek-VL2\cite{deepseek}. The architectural specifications, training protocols, and implementation details of all models are fully presented in Appendix \ref{VLM_information}.

\begin{table*}[!t]
\resizebox{\textwidth}{!}{%
\begin{tabular}{lcccccccccccccc}
\hline
\multicolumn{1}{c}{} &
  \multicolumn{6}{c|}{\textbf{Closed-source VLMs}} &
  \multicolumn{5}{c}{\textbf{Open-source VLMs}} &
   &
   &
   \\ \cline{2-12}
\multicolumn{1}{c}{\multirow{-2}{*}{\textbf{Task}}} &
  \textbf{\begin{tabular}[c]{@{}c@{}}GPT\\ 4o\end{tabular}} &
  \textbf{\begin{tabular}[c]{@{}c@{}}Gemini\\ 2.0\end{tabular}} &
  \textbf{\begin{tabular}[c]{@{}c@{}}Gemini\\ 1.5\end{tabular}} &
  \textbf{\begin{tabular}[c]{@{}c@{}}Claude\\ 3.5\end{tabular}} &
  \textbf{GLM} &
  \multicolumn{1}{c|}{\textbf{Yi}} &
  \textbf{Qwen} &
  \textbf{\begin{tabular}[c]{@{}c@{}}Intern\\ VL2\end{tabular}} &
  \textbf{PixTral} &
  \textbf{QVQ} &
  \textbf{\begin{tabular}[c]{@{}c@{}}Deep\\ Seek\end{tabular}} &
  \multirow{-2}{*}{\textbf{\begin{tabular}[c]{@{}c@{}}Human\\ Eval.\end{tabular}}} &
  \multirow{-2}{*}{\textbf{\begin{tabular}[c]{@{}c@{}}Int.\\ Avg\end{tabular}}} &
  \multirow{-2}{*}{\textbf{\begin{tabular}[c]{@{}c@{}}Sep.\\ Avg\end{tabular}}} \\ \hline
\textbf{\cellcolor[HTML]{F5F5F5}Trans.} &
  {\ul 27.66} &
  22.34 &
  {\ul 27.66} &
  24.47 &
  23.40 &
  25.53 &
  {\ul \textbf{32.98}} &
  14.89 &
  {\ul 28.72} &
  {\ul 26.60} &
  {\ul 30.85} &
  73.28 &
  25.92 &
  26.73 \\
\textbf{\cellcolor[HTML]{F5F5F5}Rot.} &
  {\ul 28.57} &
  23.21 &
  {\ul \textbf{30.36}} &
  21.43 &
  23.21 &
  23.21 &
  {\ul \textbf{30.36}} &
  {\ul \textbf{30.36}} &
  19.64 &
  25.00 &
  25.00 &
  77.30 &
  25.49 &
  28.79 \\
\textbf{\cellcolor[HTML]{F5F5F5}Comb.} &
  {\ul 30.00} &
  {\ul 36.67} &
  23.33 &
  16.67 &
  23.33 &
  20.00 &
  {\ul \textbf{43.33}} &
  {\ul 40.00} &
  23.33 &
  {\ul 40.00} &
  {\ul 30.00} &
  75.53 &
  29.70 &
  25.83 \\ \hline
\textbf{\cellcolor[HTML]{E8F0FE}Trav.} &
  24.07 &
  {\ul \textbf{31.48}} &
  22.22 &
  {\ul 27.78} &
  {\ul \textbf{31.48}} &
  24.07 &
  22.22 &
  24.07 &
  22.22 &
  {\ul 29.63} &
  22.22 &
  69.07 &
  25.59  &
  25.00 \\
\textbf{\cellcolor[HTML]{E8F0FE}Add.\&Sub.} &
  {\ul \textbf{32.84}} &
  26.87 &
  23.88 &
  20.90 &
  {\ul 29.85} &
  {\ul 28.36} &
  {\ul 29.85} &
  23.88 &
  {\ul \textbf{32.84}} &
  16.42 &
  {\ul 31.34} &
  73.19 &
  27.00 &
  28.17 \\
\textbf{\cellcolor[HTML]{E8F0FE}B\&W} &
  23.81 &
  {\ul 34.92} &
  {\ul 26.98} &
  {\ul 33.33} &
  14.29 &
  {\ul 25.40} &
  23.81 &
  {\ul \textbf{36.51}} &
  20.63 &
  12.70 &
  17.46 &
  70.67 &
  24.53 &
  26.59 \\ \hline
\textbf{\cellcolor[HTML]{E8F5E9}Sym.} &
  {\ul \textbf{35.78}} &
  {\ul 28.44} &
  19.27 &
  16.51 &
  {\ul 34.86} &
  21.10 &
  22.94 &
  22.02 &
  {\ul 27.52} &
  22.94 &
  20.18 &
  73.61 &
  24.69 &
  25.11 \\
\textbf{\cellcolor[HTML]{E8F5E9}O\&C } &
  21.05 &
  26.32 &
  {\ul 31.58} &
  15.79 &
  {\ul 36.84} &
  {\ul 36.84} &
  {\ul 31.58} &
  {\ul \textbf{47.37}} &
  {\ul 36.84} &
  5.26 &
  {\ul 31.58} &
  73.89 &
  29.19 &
  30.26 \\
\textbf{\cellcolor[HTML]{E8F5E9}Comb.} &
  16.67 &
  16.67 &
  16.67 &
  16.67 &
  16.67 &
  16.67 &
  16.67 &
  {\ul 33.33} &
  {\ul \textbf{50.00}} &
  {\ul 33.33} &
  16.67 &
  72.50 &
  22.73 &
  14.58 \\ \hline
\textbf{\cellcolor[HTML]{FFFDE7}Line} &
  20.81 &
  {\ul \textbf{31.21}} &
  {\ul 25.43} &
  23.12 &
  {\ul 25.43} &
  {\ul 28.32} &
  21.39 &
  21.97 &
  18.50 &
  20.81 &
  {\ul 24.86} &
  64.71 &
  23.80  &
  24.06 \\
\textbf{\cellcolor[HTML]{FFFDE7}Surf.} &
  23.36 &
  22.63 &
  {\ul 27.01} &
  {\ul \textbf{28.47}} &
  {\ul 26.28} &
  18.25 &
  {\ul 27.01} &
  21.17 &
  21.90 &
  {\ul 27.01} &
  {\ul 26.28} &
  71.50 &
  24.49  &
  26.55 \\
\textbf{\cellcolor[HTML]{FFFDE7}Point} &
  {\ul 28.79} &
  15.15 &
  16.67 &
  21.21 &
  {\ul 28.79} &
  16.67 &
  {\ul 22.73} &
  {\ul \textbf{30.30}} &
  {\ul 24.24} &
  21.21 &
  19.70 &
  65.24 &
  22.31 &
  25.00 \\
\textbf{\cellcolor[HTML]{FFFDE7}Elem.} &
  21.28 &
  {\ul 27.66} &
  23.40 &
  {\ul 24.47} &
  {\ul 25.53} &
  {\ul 24.47} &
  {\ul \textbf{30.85}} &
  18.09 &
  21.28 &
  17.02 &
  {\ul 29.79} &
  63.97 &
  23.98 &
  25.80 \\
\textbf{\cellcolor[HTML]{FFFDE7}Comb.} &
  20.00 &
  22.00 &
  16.00 &
  {\ul 30.00} &
  {\ul 34.00} &
  {\ul \textbf{38.00}} &
  20.00 &
  {\ul 26.00} &
  {\ul 26.00} &
  24.00 &
  {\ul 26.00} &
  64.30 &
  25.64 &
  24.25 \\ \hline
\textbf{\cellcolor[HTML]{FCF2E3}Cube} &
  {\ul 27.52} &
  {\ul 25.69} &
  23.85 &
  {\ul 28.44} &
  {\ul 27.52} &
  {\ul 28.44} &
  21.10 &
  {\ul 27.52} &
  {\ul \textbf{30.28}} &
  21.10 &
  20.18 &
  64.31 &
  25.60 &
  25.11 \\
\textbf{\cellcolor[HTML]{FCF2E3}3D Mos.} &
  17.39 &
  {\ul \textbf{36.96}} &
  {\ul 30.43} &
  15.22 &
  {\ul 32.61} &
  23.91 &
  {\ul 26.09} &
  19.57 &
  23.91 &
  19.57 &
  17.39 &
  54.46 &
  23.91 &
  25.54 \\
\textbf{\cellcolor[HTML]{FCF2E3}Poly.} &
  23.53 &
  17.65 &
  {\ul 29.41} &
  {\ul \textbf{35.29}} &
  17.65 &
  23.53 &
  {\ul \textbf{35.29}} &
  {\ul \textbf{35.29}} &
  {\ul 29.41} &
  17.65 &
  {\ul 29.41} &
  60.18 &
  26.74  &
  22.06 \\
\textbf{\cellcolor[HTML]{FCF2E3}3-View} &
  {\ul \textbf{32.50}} &
  25.00 &
  {\ul 30.00} &
  25.00 &
  25.00 &
  25.00 &
  {\ul 30.00} &
  25.00 &
  25.00 &
  25.00 &
  17.50 &
  68.50 &
  25.91 &
  23.75 \\
\textbf{\cellcolor[HTML]{FCF2E3}Sec.View} &
  25.71 &
  17.14 &
  22.86 &
  22.86 &
  {\ul \textbf{42.86}} &
  {\ul 28.57} &
  {\ul 31.43} &
  {\ul 31.43} &
  {\ul 34.29} &
  14.29 &
  {\ul 28.57} &
  55.69 &
  27.27 &
  28.93 \\
\textbf{\cellcolor[HTML]{FCF2E3}Q\&T} &
  {\ul \textbf{40.00}} &
  {\ul 30.00} &
  20.00 &
  20.00 &
  {\ul 30.00} &
  20.00 &
  {\ul \textbf{40.00}} &
  10.00 &
  {\ul 30.00} &
  20.00 &
  10.00 &
  68.80 &
  24.55 &
  32.50 \\ \hline
\textbf{\cellcolor[HTML]{F3E5F5}2D Comp.} &
  {\ul 29.03} &
  {\ul 25.81} &
  19.35 &
  {\ul 32.26} &
  19.35 &
  19.35 &
  19.35 &
  {\ul \textbf{38.71}} &
  22.58 &
  {\ul 32.26} &
  22.58 &
  70.58 &
  25.51 &
  25.00 \\
\textbf{\cellcolor[HTML]{F3E5F5}InterFig.} &
  {\ul \textbf{40.00}} &
  {\ul 30.00} &
  27.50 &
  22.50 &
  {\ul 30.00} &
  {\ul 32.50} &
  27.50 &
  22.50 &
  {\ul 32.50} &
  17.50 &
  22.50 &
  68.18 &
  27.73 &
  20.94 \\ \hline
\textbf{\cellcolor[HTML]{FBF4BD}AlphaNum.} &
  29.63 &
  {\ul 33.33} &
  {\ul 33.33} &
  {\ul \textbf{40.74}} &
  25.93 &
  {\ul 33.33} &
  {\ul 37.04} &
  22.22 &
  29.63 &
  {\ul \textbf{40.74}} &
  18.52 &
  54.96 &
  31.31 &
  32.41 \\ \hline
\textbf{\cellcolor[HTML]{D0DFE3}B\&W } &
  {\ul 28.13} &
  21.88 &
  {\ul \textbf{31.25}} &
  21.88 &
  21.88 &
  21.88 &
  {\ul 25.00} &
  18.75 &
  {\ul \textbf{31.25}} &
  12.50 &
  {\ul 25.00} &
  68.47 &
  23.58  &
  23.83 \\ \hline
\textbf{\cellcolor[HTML]{FCD6D2}Misc.} &
  {\ul 35.29} &
  {\ul 38.24} &
  {\ul \textbf{47.06}} &
  {\ul 44.12} &
  {\ul 38.24} &
  14.71 &
  23.53 &
  20.59 &
  {\ul 41.18} &
  32.35 &
  {\ul 41.18} &
  60.44 &
  34.22 &
  30.88 \\ \hline
\textbf{\cellcolor[HTML]{E4E9DB}Mensa-v1} &
  17.14 &
  {\ul \textbf{34.29}} &
  {\ul 20.00} &
  {\ul 31.43} &
  11.43 &
  5.71 &
  17.14 &
  8.57 &
  11.43 &
  {\ul 31.43} &
  {\ul 28.57} &
  80.00 &
  19.74 &
  16.79 \\
\textbf{\cellcolor[HTML]{E4E9DB}Mensa-v2} &
  {\ul 20.51} &
  {\ul 23.08} &
  12.82 &
  15.38 &
  {\ul 20.51} &
  15.38 &
  {\ul 20.51} &
  10.26 &
  15.38 &
  {\ul 20.51} &
  {\ul \textbf{25.64}} &
  84.62 &
  18.18 &
  14.10 \\ \hline
\textbf{\cellcolor[HTML]{E0F7FA}Raven-v1} &
  25.00 &
  {\ul 30.00} &
  {\ul \textbf{37.50}} &
  {\ul \textbf{37.50}} &
  25.00 &
  17.50 &
  22.50 &
  25.00 &
  25.00 &
  {\ul 30.00} &
  10.00 &
  80.00 &
  25.91 &
  16.88 \\
\textbf{\cellcolor[HTML]{E0F7FA}Raven-v2} &
  25.00 &
  {\ul 31.67} &
  21.67 &
  {\ul \textbf{43.33}} &
  {\ul 35.00} &
  23.33 &
  21.67 &
  16.67 &
  {\ul 28.33} &
  23.33 &
  {\ul 30.00} &
  95.00 &
  27.27 &
  19.38 \\ \hline
\multicolumn{1}{c}{\textbf{All Avg.}} &
  26.22 &
  \textbf{27.22} &
  25.05 &
  25.85 &
  24.49 &
  23.93 &
  \textbf{25.73} &
  23.62 &
  25.17 &
  22.88 &
  24.55 &
  69.76 &
  24.97 &
  25.23 \\ 
  \hline
\end{tabular}%
}
\tablefootnote{\scriptsize The \textbf{bolded values} represent the highest accuracy results and the {\ul underlined values} represent results above the average accuracy for each task.} 
\caption{The accuracy of different VLMs on ReasonBench.}


\label{tab:all}
\end{table*}

\subsection{Benchmark evaluation}
We evaluated all baseline VLMs on ReasonBench and summarized the results in Table \ref{tab:all}. Human Baseline indicates the accuracy of human completing the ReasonBench questions. Int. Avg refers to inputting the integrated graphic into VLMs and Sep. Avg refers to separated graphical input them into VLMs sequentially.All Avg. represents the average accuracy for each column. Additionally, we provide the results of the separated format in the Table \ref{tab:cut} in Appendix. Our findings are summarized as follows:

\textbf{There are significant differences in the performance of different VLMs in various graphics reasoning tasks.} Overall, open-source models outperformed closed-source models in this category. See Table \ref{fig:result} for details.

\begin{table}[!t]
    \centering
    \vspace*{-10pt} 
    \includegraphics[width=\linewidth]{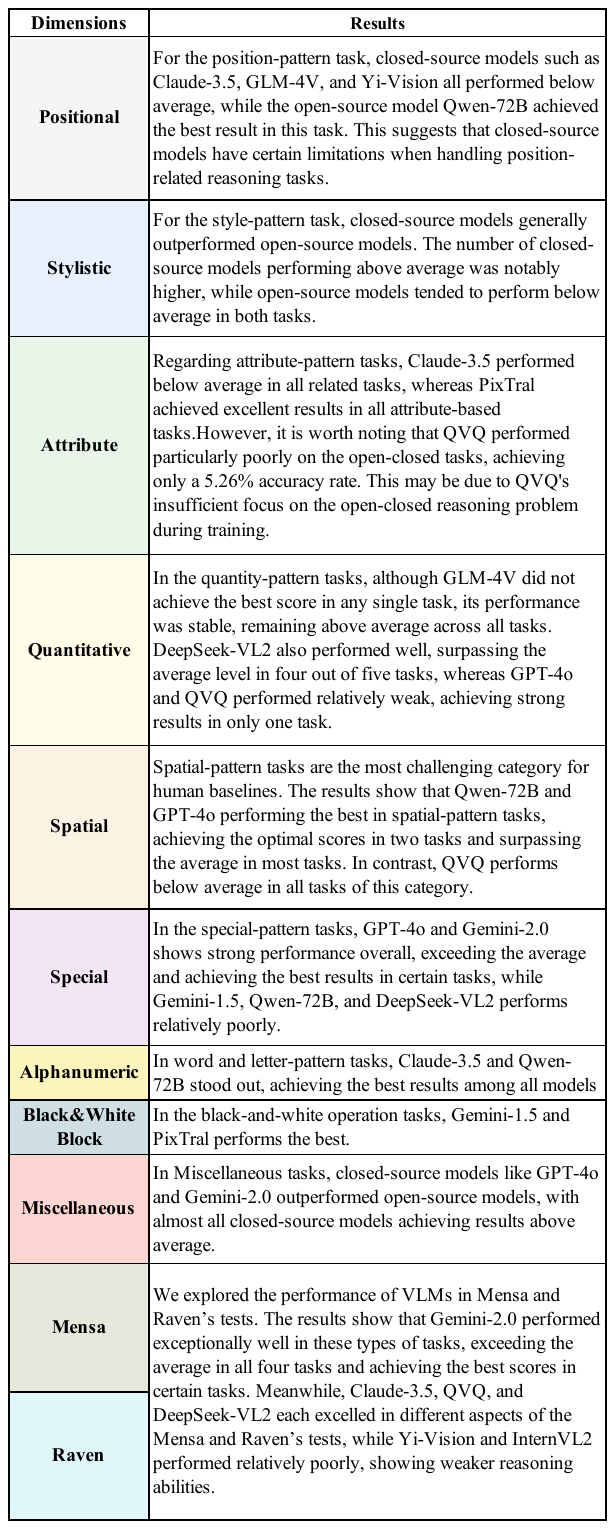}
    \caption{An overview of the VLMs' evaluation results under different Dimensions. \textbf{Dimensions} correspond to those in Table \ref{fig:dataset} and \textbf{Result} reflect the experiment details in Table \ref{tab:all}.}
    \label{fig:result}
\end{table}

\textbf{The overall answer accuracy of VLMs remains consistently around 25\%, with even the best-performing model, Gemini-2.0, achieving only 27\%.} This result is surprising, as even a model that randomly selects answers would achieve a 25\% accuracy rate among four options. In terms of overall performance, there is a significant disparity between the capabilities of VLMs and humans. Specifically, even the best-performing model falls far behind the lowest human baseline, with a gap of up to 13\%, while the difference with the highest-performing human baseline exceeds 50\%. These findings suggest that current VLMs are not yet capable of independently reasoning and accurately selecting answers from complex graphics. Therefore, the ReasonBench benchmark is crucial in highlighting the limitations of VLMs in complex graphical reasoning tasks. In other words, at least in the context of real-world complex graphical reasoning tasks, the performance of VLMs has not yet reached a level of reliability.

\textbf{Gemini-2.0 and Qwen-72B-VL represent the best-performing closed-source and open-source models, respectively. }From the data in the table, it is evident that Qwen-72B performs slightly worse than Gemini-2.0 in the Mensa and Raven tests, with Qwen-72B lagging behind in each task. Although models such as Yi-vision, QVQ-72B, and Deepseek-VL are specifically optimized for graphic analysis tasks, they do not show a more significant advantage over Qwen in overall performance. Notably, while QVQ outperforms Qwen-72B in various metrics in official showcases, it ranks lower in our benchmark tests. In the overall assessment, open-source models generally outperform closed-source models across most tasks. However, it is noteworthy that Qwen and pixtral have surpassed GLM and Gemini-1.5, achieving leading positions in multiple tasks. Despite this, when comparing across different task categories, closed-source models still lead in a greater number of tasks, including cases of shared first place. The recently popular Deepseek, we also tested its VL2 version, which only achieved a leading position in one task.

\textbf{Overall, the difference in performance between the integrated and separated approaches for VLMs is minimal. In general, the integrated approach achieves an accuracy of 25.26\%, while the separated approach reaches an accuracy of 25.23\%.} According to the comparison data in Table 1 and Appendix \ref{Sep_table} Table \ref{tab:cut}, in tasks involving positional dimensions and style dimensions, the separated questioning method outperforms the integrated method for models like Gemini-1.5, Claude-3.5, and Yi-vision. However, Qwen shows superior performance in positional dimension tasks when using the integrated method, significantly outperforming the separated method. For attribute dimension tasks, Gemini-2.0 and Claude-3.5 perform better with the separated method than with the integrated method. In numerical dimension tasks, Gemini-1.5 and InternVL perform better with the separated method. For spatial dimension tasks, Gemini-1.5 and Qwen show better results with the integrated method compared to the separated method. In black-and-white arithmetic and comprehensive dimension tasks, the integrated method outperforms the separated method for Gemini-2.0, Gemini-1.5, and Claude-3.5.
Among the 232 sets of data compared, 38 sets were completely consistent.

\section{Improvement Methods}
We propose a dual optimization solution, which is suitable for closed-source and open-source models, Diagrammatic Chain-of-Thought (DiaCoT) and fine-tuning strategy (ReasonTune). We randomly select 200 questions from the dataset as the validation set for the improvement method. The validation set meets all the previous conditions. The purpose is to be able to make horizontal comparisons in ReasonTune, so in the improvement experiment, we use the validation set uniformly.

\subsection{DiaCoT}
To enhance the generalization of the method across both open-source and closed-source models, we first consider prompt engineering and explore effective problem-solving strategies based on Chain-of-Thought\cite{COT}. While strategies such as visual perception, inductive reasoning, and deductive reasoning are highly effective in addressing simple single-element and two-element problems\cite{PuzzleVQA_ACL2024}, their performance is limited when dealing with more complex multi-element problems. Therefore, we propose a novel reasoning method—\textbf{DiaCoT}.

\begin{figure}[htbp]
\vspace{0.2cm}  
  \centering
  \includegraphics[width=\linewidth]{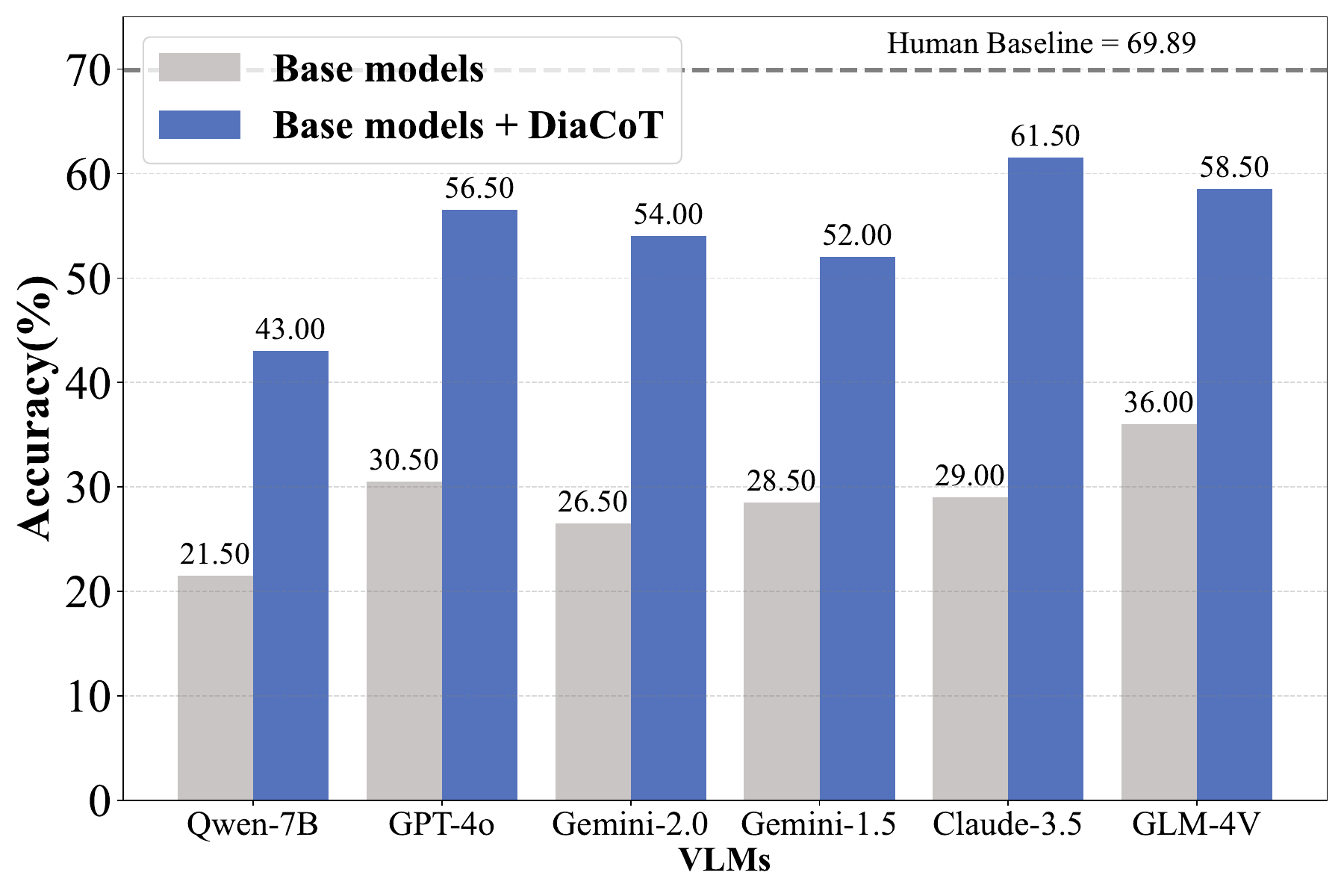}
  \caption{Comparison between Base models and DiaCoT method.}
  \label{fig:DiaCoT_base}
\end{figure}

Specifically, in designing prompts, we guide the VLMs to explain the content of the question and options individually and decompose multi-element diagrams layer by layer from a graphical perspective. In addition, DiaCoT integrates the contextual information of the problem environment, providing the model with the necessary context, defining the problem scope, and enhancing the model's abstract reasoning capabilities through the prompts. Detailed implementations and examples can be found in Appendix \ref{format}.

In experimental testing, we compare DiaCoT with the baseline methods, As shown in Figure \ref{fig:DiaCoT_base}. All models show a apparent improvement in accuracy with an average increase of 25.58\% after applying DiaCoT and narrowing the gap with human-level performance. Although the accuracy has not yet achieved human-level precision, it has successfully exceeded the limitation where models typically perform below 25\% accuracy or rely on random guessing.

\subsection{ReasonTune}
\begin{figure}[htbp]
\vspace{0.2cm}
  \includegraphics[width=0.48\linewidth]{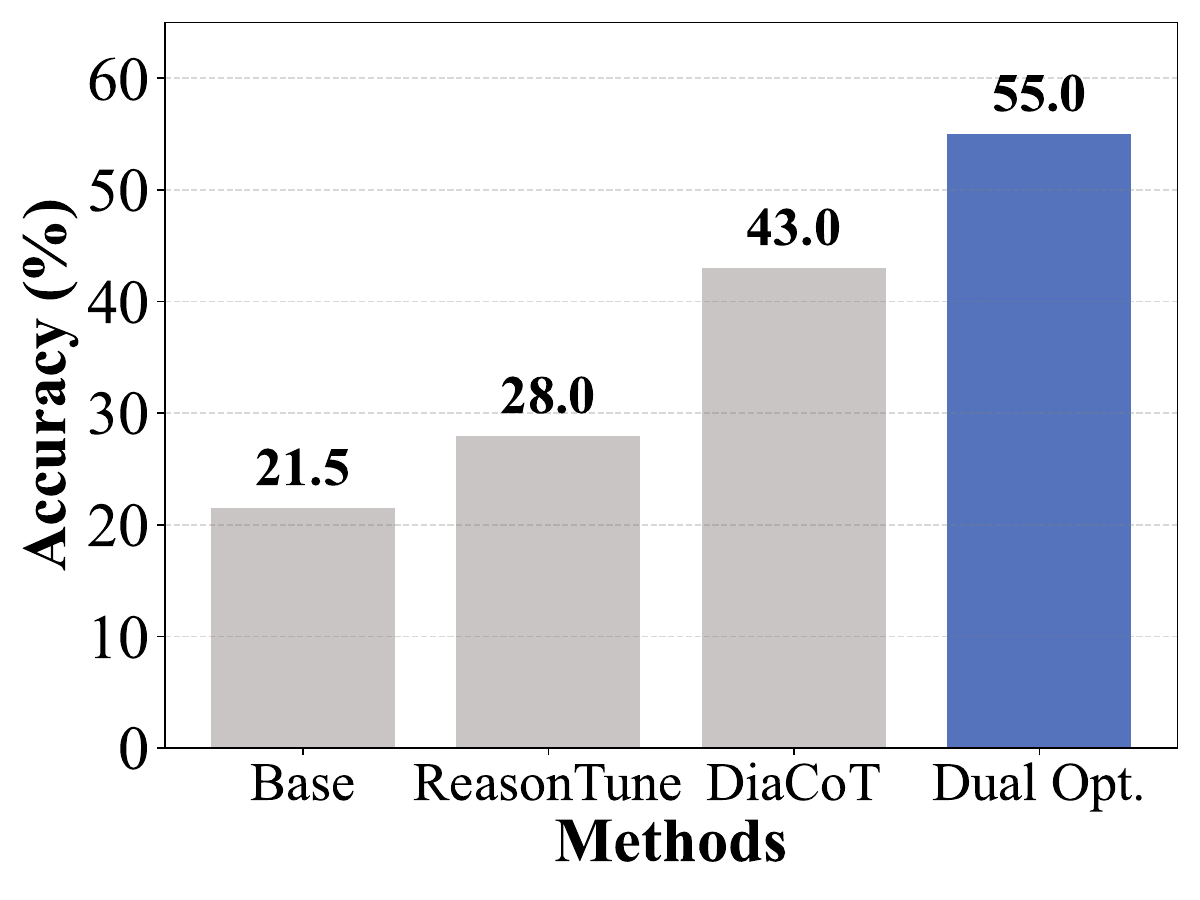} \hfill
  \includegraphics[width=0.48\linewidth]{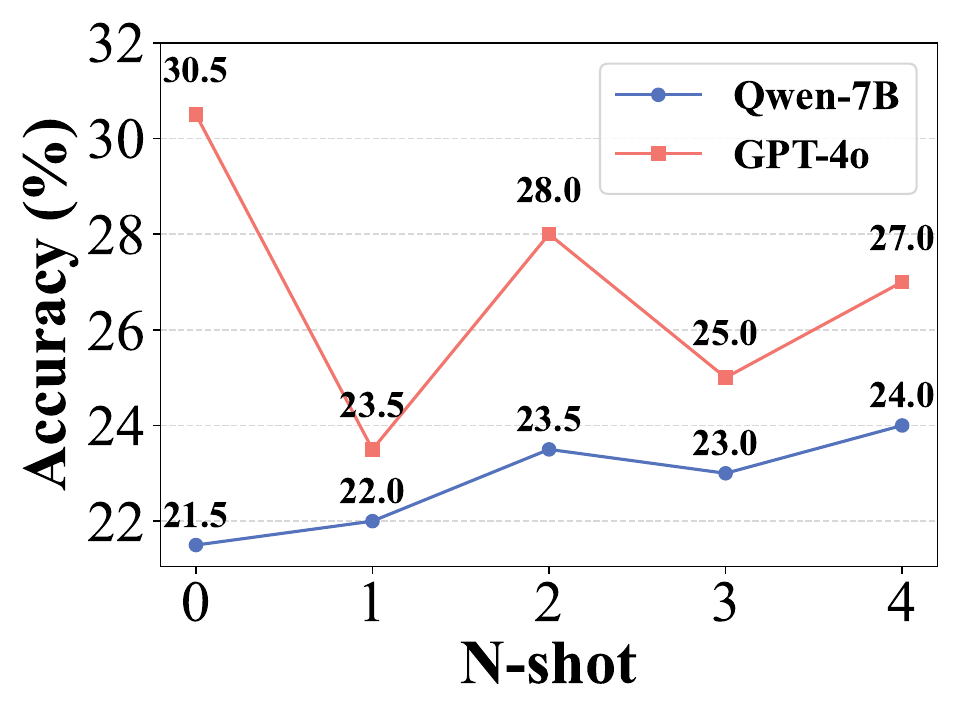}
  \caption {Left: Ablation results of Dual Optimization. Right: Analysis of the impact of few-shot demonstrations on model performance in complex graph reasoning.}
  \label{fig4_a}
\end{figure}

We selected Qwen-7B\cite{bai2023qwen} as the open source model for ReasonTune optimization (due to its poor performance), and fine-tuned and optimized it using datasets other than the validation set. As shown in Figure \ref{fig4_a} left, the accuracy of Qwen-7B is only 21.5\%. Through ablation experiments, we verified the synergy of dual optimization: (1) ReasonTune on the base model improved the accuracy by 6.5\%. (2) The introduction of the DiaCoT method improved the performance by 21.5\%. (3) The proposed dual optimization framework achieved an overall improvement of 33.5\%. This gradual performance improvement not only verifies the effectiveness of our optimization, but also provides a new optimization paradigm for models that handle complex graph reasoning.



\subsection{Effect of Few-Shot Demonstrations}
\label{few_shot}
In the evaluation, we use zero-shot demonstrations to evaluate how VLM handles complex graph reasoning tasks. In addition, we explore the performance of the model when using few-shot demonstrations. To ensure the diversity of demonstrations, we randomly select questions of different dimensions from the database for each few-shot demonstration.

The results show that the performance improves with the number of demonstrations. However, we find that for GPT-4o, the performance does not reach a peak even with the largest number of demonstrations. This suggests that the relationship between reasoning ability and contextual learning may be proportional but nonlinear when dealing with similar complex graph reasoning tasks in the real world. Nevertheless, we still prove that few-shot demonstrations are a reliable method, based on the performance of Qwen-7B, which highlights the effectiveness of this method in complex graph reasoning tasks.

\section{Conclusion}
In order to fairly and effectively evaluate the reasoning ability of VLMs on complex graph reasoning task, we develop a new benchmark named ReasonBench, which contains 1,613 questions across 11 cognitive dimensions and 29 task types. We select 11 mainstream VLMs and evaluate their performances on ReasonBench. The results reveal that the existing models have limitations: even the best-performing model can only achieve 27\% accuracy, far below the human baseline of 68.7\%. This gap highlights that the ability of VLMs to reason about complex graphs needs to be improved. To address these issues, we propose a dual optimization framework. The DiaCoT method enhances the interpretability of reasoning by decomposing layers, while ReasonTune enhances reasoning task adaptability through training. Experimental results show that the overall reasoning accuracy is improved by 33.5\%, verifying the effectiveness of our framework. Our study lays a foundation for developing VLMs with complex graph reasoning.

\section{Limitation}
In our work, training the model requires a large amount of data, which may exhaust the resources of the validation set. Therefore, no comprehensive validation is performed on the entire dataset during the model optimization stage. But this conservative approach ensures the rigor of the method validation. As a compensation mechanism, we systematically collect response data for each model on a limited validation set and establish a relative performance evaluation framework through multi-level horizontal comparisons to ensure the effectiveness and reliability of the evaluation. At the same time, we set up few-shot experiments in Section \ref{few_shot}, which show certain regularities.


\bibliography{acl}

\clearpage  
\appendix

\section{Appendix} \label{appendix-A}

\subsection{Multiple Choice Question (MCQ) Distribution}
\label{MCQ_distribution}
\begin{figure*}[htbp]
  \includegraphics[width=0.32\linewidth]{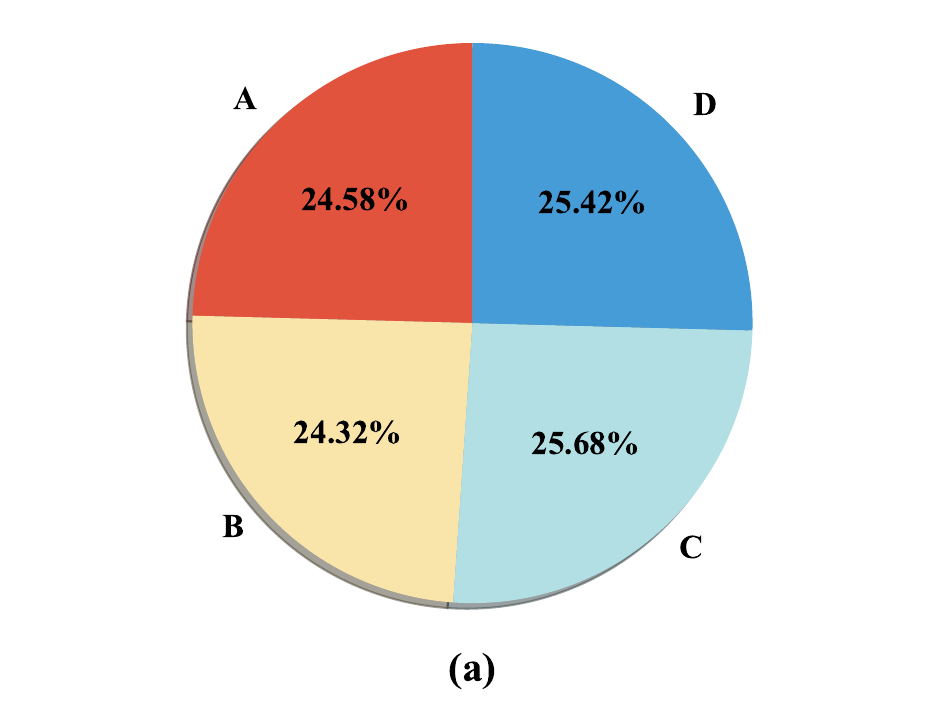} 
  \includegraphics[width=0.32\linewidth]{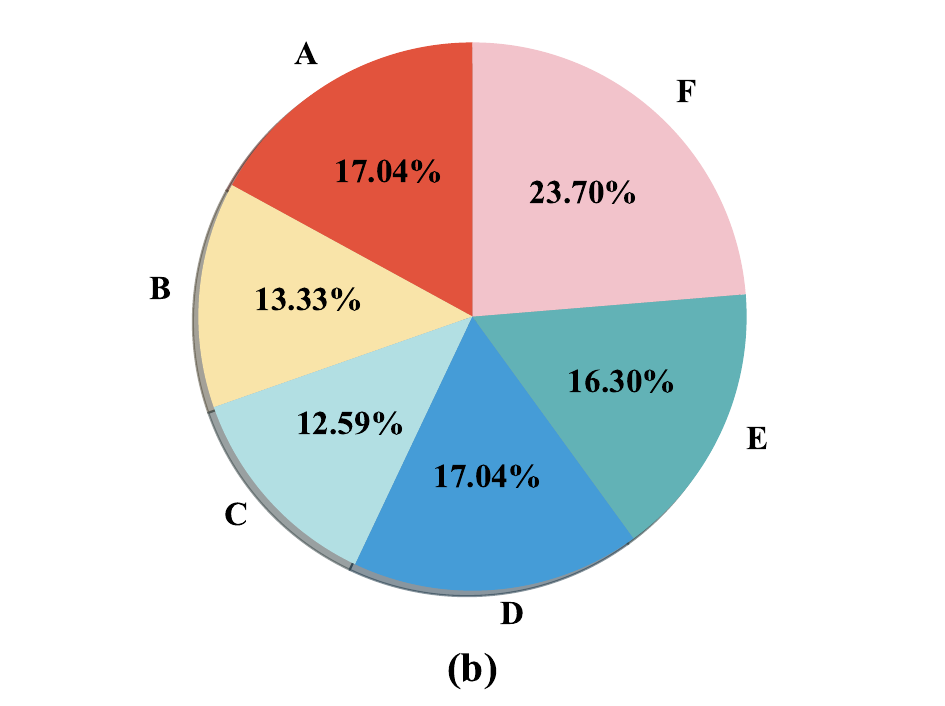} 
  \includegraphics[width=0.32\linewidth]{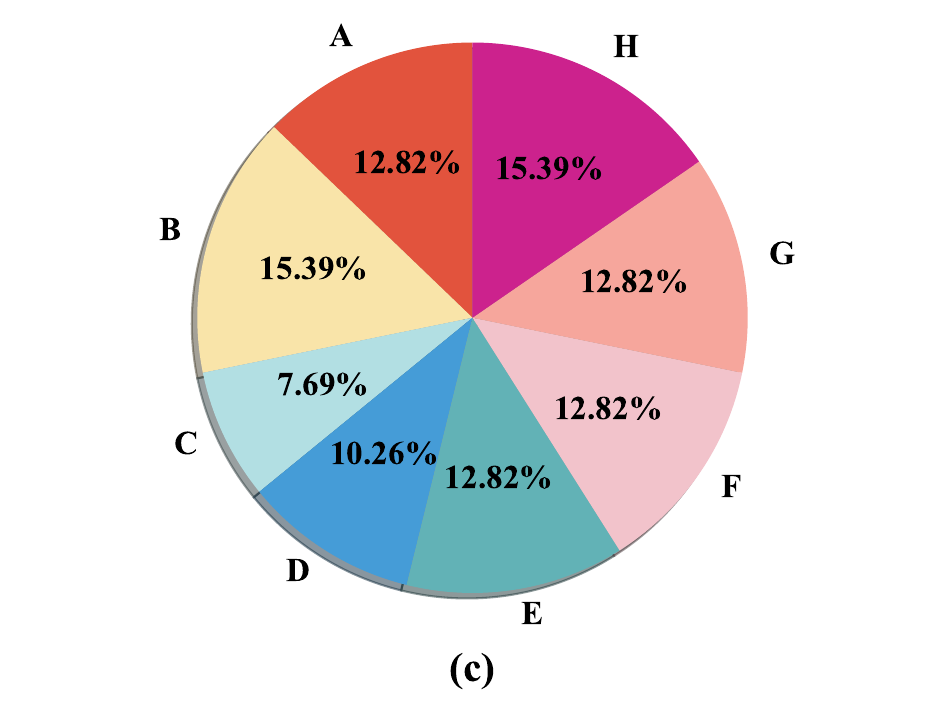}
  \caption {Pie charts of the MCQ distribution. These three charts represent the probability distribution of selecting each choice when the number of choices is 4, 6, and 8, respectively. We label the probability of selecting different choice within the sectors and use different colors for distinction.}
  \label{fig:MCQ}
\end{figure*}

The distribution of MCQ is shown in Figure \ref{fig:MCQ}. Here, Figure \ref{fig:MCQ}(a) represents the probability distribution of selecting each option when there are four choices, Figure \ref{fig:MCQ}(b) represents the probability distribution when there are six choices, and Figure \ref{fig:MCQ}(c) shows the probability distribution when there are eight choices. It can be observed that when the number of choices is four or eight, the probabilities of selecting each choice are very close. When there are six, the probability of selecting choice F is slightly higher. Overall, as the number of choice increases, the probabilities of selecting each one remain fairly balanced. And there is no significant preference for any specific choice or a few ones.

\begin{table*}[htbp]
\centering
\includegraphics[width=\linewidth]{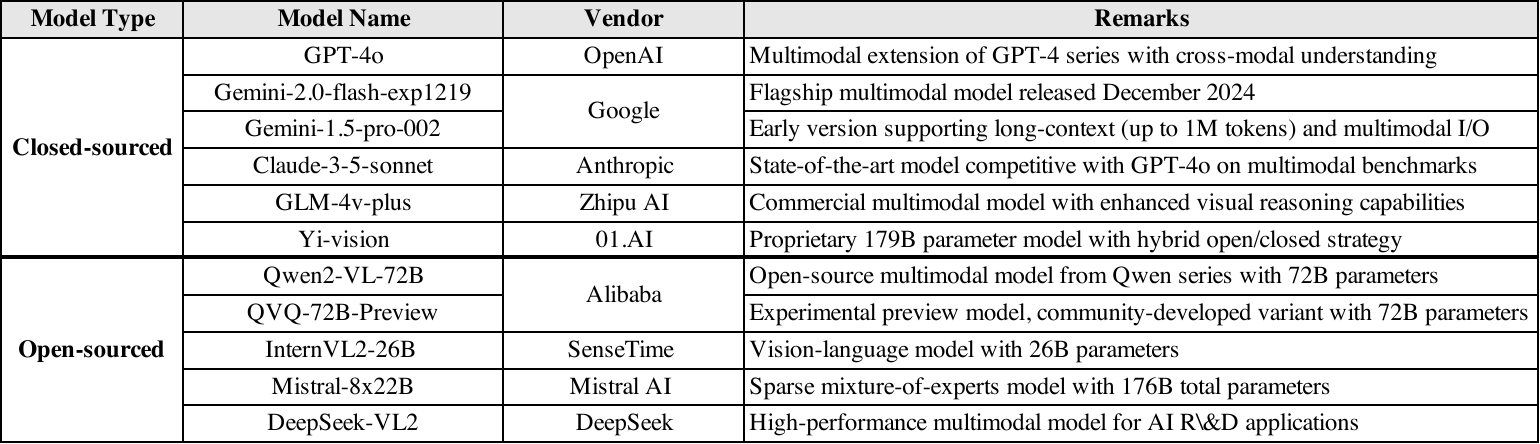} 
\caption{Model Information Comparison. We classify the models that used in our paper based on whether they are open-source and label the vendor of each one along with a brief remark.}
\label{tab:model_comparison}
\end{table*}


\subsection{Pass@k}
\label{pass@k}
To evaluate the graphical reasoning capabilities of VLMs, we introduce Pass@k metric from the code generation domain. We choose \textbf{Pass@1} because it directly reflects the model's deterministic understanding of the problem. A high Pass@1 rate indicates that the model consistently generates the correct solution, rather than relying on "luck", while a high Pass@100 may rely on diversity in generation to mask the model's deficiencies.

\vspace{-5pt}
\subsection{VLMs}
\label{VLM_information}
Model version\&company information, see Table \ref{tab:model_comparison}.

\vspace{-5pt}
\subsection{Separation format evaluation results}
\label{Sep_table}
Due to the fact that some VLM API services were interrupted due to force majeure and GLM does not support uploading 7 graphics at the same time, we conducted the test on 8 VLMs. See Table \ref{tab:cut} for the evaluation results of the separation format.

\begin{table*}[htbp]
\tiny
\resizebox{\textwidth}{!}{%
\begin{tabular}{lcccccccc}
\hline
\multicolumn{1}{c}{\textbf{Task}} &
  \textbf{\begin{tabular}[c]{@{}c@{}}GPT\\ 4o\end{tabular}} &
  \textbf{\begin{tabular}[c]{@{}c@{}}Gemini\\ 2.0\end{tabular}} &
  \textbf{\begin{tabular}[c]{@{}c@{}}Gemini\\ 1.5\end{tabular}} &
  \textbf{\begin{tabular}[c]{@{}c@{}}Claude\\ 3.5\end{tabular}} &
  \textbf{\begin{tabular}[c]{@{}c@{}}GLM\\ 4V\end{tabular}} &
  \textbf{\begin{tabular}[c]{@{}c@{}}Intern\\ VL2\end{tabular}} &
  \textbf{\begin{tabular}[c]{@{}c@{}}Qwen\\ 7B\end{tabular}} &
  \textbf{\begin{tabular}[c]{@{}c@{}}Yi\\ Vision\end{tabular}} \\ \hline
\textbf{\cellcolor[HTML]{F5F5F5}Trans.}     & 29.79 & 27.66 & 27.66 & 25.53 & 28.72 & 27.66 & 21.28 & 25.53 \\
\textbf{\cellcolor[HTML]{F5F5F5}Rot.}        & 30.36 & 23.21 & 33.93 & 33.93 & 23.21 & 23.21 & 26.79 & 35.71 \\
\textbf{\cellcolor[HTML]{F5F5F5}Comb.} & 23.33 & 30.00 & 26.67 & 26.67 & 20.00 & 33.33 & 20.00 & 26.67 \\
\textbf{\cellcolor[HTML]{E8F0FE}Trav.}        & 27.78 & 16.67 & 25.93 & 35.19 & 22.22 & 14.81 & 31.48 & 25.93 \\
\textbf{\cellcolor[HTML]{E8F0FE}Add.\&Sub.}               & 28.36 & 37.31 & 25.37 & 28.36 & 22.39 & 20.90 & 31.34 & 31.34 \\
\textbf{\cellcolor[HTML]{E8F0FE}B\&W}           & 26.98 & 28.57 & 31.75 & 19.05 & 31.75 & 25.40 & 30.16 & 19.05 \\
\textbf{\cellcolor[HTML]{E8F5E9}Sym.}        & 27.52 & 33.03 & 23.85 & 31.19 & 14.68 & 23.85 & 24.77 & 22.02 \\
\textbf{\cellcolor[HTML]{E8F5E9}O\&C }          & 36.84 & 31.58 & 31.58 & 42.11 & 31.58 & 15.79 & 26.32 & 26.32 \\
\textbf{\cellcolor[HTML]{E8F5E9}Comb.}        & 33.33 & 16.67 & 0.00  & 16.67 & 33.33 & 0.00  & 0.00  & 16.67 \\
\textbf{\cellcolor[HTML]{FFFDE7}Line}            & 24.28 & 24.86 & 23.12 & 19.65 & 25.43 & 23.12 & 26.01 & 26.01 \\
\textbf{\cellcolor[HTML]{FFFDE7}Surf.}         & 22.63 & 26.28 & 30.66 & 28.47 & 26.28 & 24.82 & 27.01 & 26.28 \\
\textbf{\cellcolor[HTML]{FFFDE7}Point}           & 27.27 & 21.21 & 24.24 & 24.24 & 22.73 & 30.30 & 28.79 & 21.21 \\
\textbf{\cellcolor[HTML]{FFFDE7}Elem.}         & 23.40 & 27.66 & 30.85 & 22.34 & 28.72 & 25.53 & 22.34 & 25.53 \\
\textbf{\cellcolor[HTML]{FFFDE7}Comb.}        & 28.00 & 22.00 & 30.00 & 18.00 & 26.00 & 32.00 & 18.00 & 20.00 \\
\textbf{\cellcolor[HTML]{FCF2E3}Cube}       & 26.61 & 19.27 & 24.77 & 22.02 & 31.19 & 27.52 & 28.44 & 21.10 \\
\textbf{\cellcolor[HTML]{FCF2E3}3D Mos.}              & 23.91 & 28.26 & 26.09 & 21.74 & 28.26 & 26.09 & 26.09 & 23.91 \\
\textbf{\cellcolor[HTML]{FCF2E3}Poly.}                    & 23.53 & 52.94 & 17.65 & 5.88  & 17.65 & 17.65 & 17.65 & 23.53 \\
\textbf{\cellcolor[HTML]{FCF2E3}3-View} & 15.00 & 20.00 & 20.00 & 22.50 & 25.00 & 27.50 & 20.00 & 40.00 \\
\textbf{\cellcolor[HTML]{FCF2E3}Sec.View}         & 22.86 & 37.14 & 20.00 & 31.43 & 17.14 & 28.57 & 25.71 & 48.57 \\
\textbf{\cellcolor[HTML]{FCF2E3}Q\&T}                  & 30.00 & 60.00 & 20.00 & 30.00 & 30.00 & 10.00 & 30.00 & 50.00 \\
\textbf{\cellcolor[HTML]{F3E5F5}2D Comp.}                  & 29.03 & 25.81 & 19.35 & 35.48 & 22.58 & 22.58 & 16.13 & 29.03 \\
\textbf{\cellcolor[HTML]{F3E5F5}InterFig.}                  & 20.00 & 20.00 & 22.50 & 17.50 & 7.50  & 30.00 & 27.50 & 22.50 \\
\textbf{\cellcolor[HTML]{FBF4BD}AlphaNum.}               & 29.63 & 40.74 & 33.33 & 22.22 & 29.63 & 33.33 & 29.63 & 40.74 \\
\textbf{\cellcolor[HTML]{D0DFE3}B\&W }           & 28.13 & 18.75 & 28.13 & 18.75 & 18.75 & 31.25 & 25.00 & 21.88 \\
\textbf{\cellcolor[HTML]{FCD6D2}Misc.}                            & 32.35 & 29.41 & 26.47 & 38.24 & 38.24 & 29.41 & 23.53 & 29.41 \\
\textbf{\cellcolor[HTML]{E4E9DB}Mensa-v1}                         & 20.00 & 25.71 & 17.14 & 17.14 & /       & 20.00 & 20.00 & 14.29 \\
\textbf{\cellcolor[HTML]{E4E9DB}Mensa-v2}                         & 17.95 & 20.51 & 25.64 & 10.26 & /       & 10.26 & 12.82 & 15.38 \\
\textbf{\cellcolor[HTML]{E0F7FA}Raven-v1}                        & 30.00 & 32.50 & 22.50 & 12.50 & /       & 10.00 & 22.50 & 5.00  \\
\textbf{\cellcolor[HTML]{E0F7FA}Raven-v2}                        & 38.33 & 30.00 & 30.00 & 16.67 & /       & 13.33 & 13.33 & 13.33 \\ \hline
\textbf{All Avg.}                                & 26.29 & 26.91 & 26.16 & 24.12 & 24.88 & 24.05 & 24.55 & 24.86 \\ \hline
\end{tabular}%
}
\caption{The evaluation results in separated format. We omit the percentage sign(\%). Each Task corresponds to the one in Table \ref{tab:all}}
\label{tab:cut}
\end{table*}

\section{Example}
\label{Appendix-B}
\vspace{-5pt} 

\subsection{Format and DiaCoT Example}
\begin{figure*}[htbp]
  \centering
  \includegraphics[width=\textwidth]{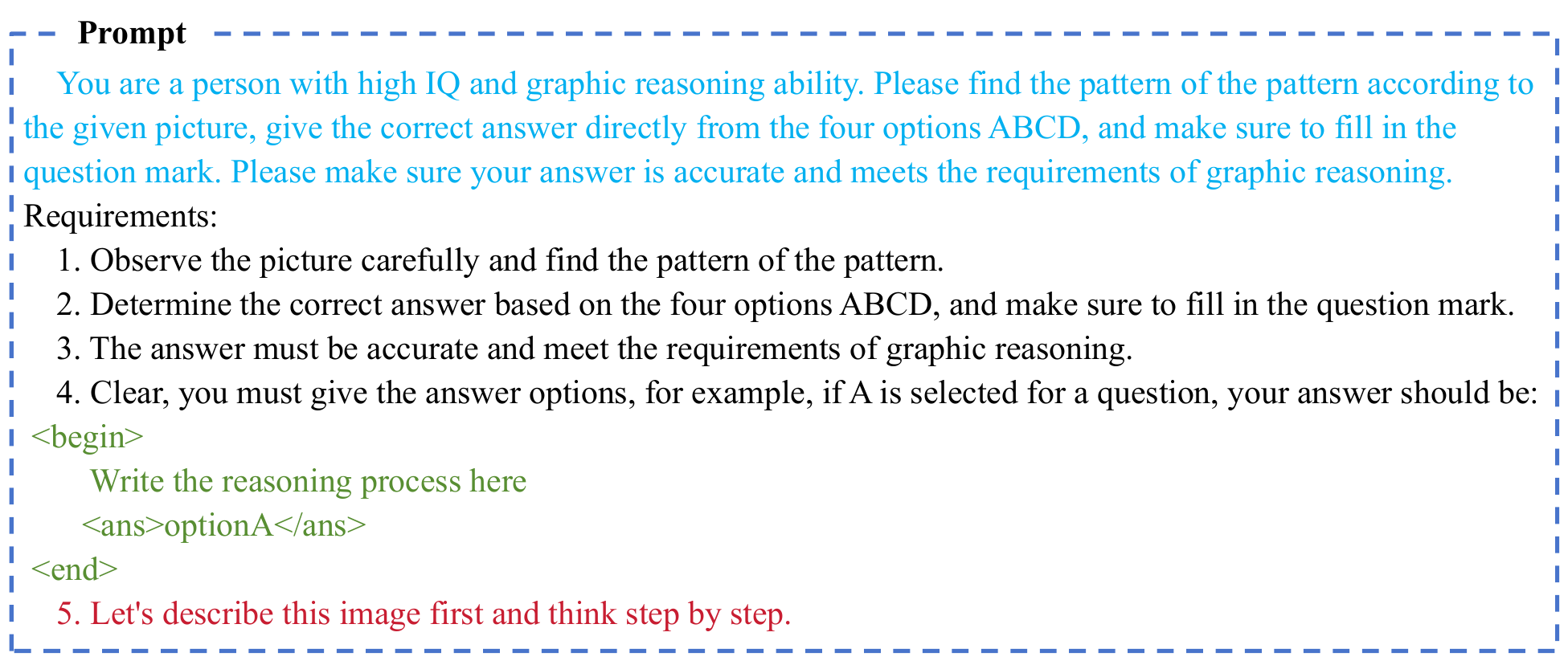}
  \caption{The prompt words of the integrated format. The blue part represents the system prompt words; the black part represents the requirements; the green part is the fixed model answer format; the red part uses the COT method.}
  \label{fig:prompt_int}
\end{figure*}

\begin{figure*}[htbp]
  \centering
  \includegraphics[width=\textwidth]{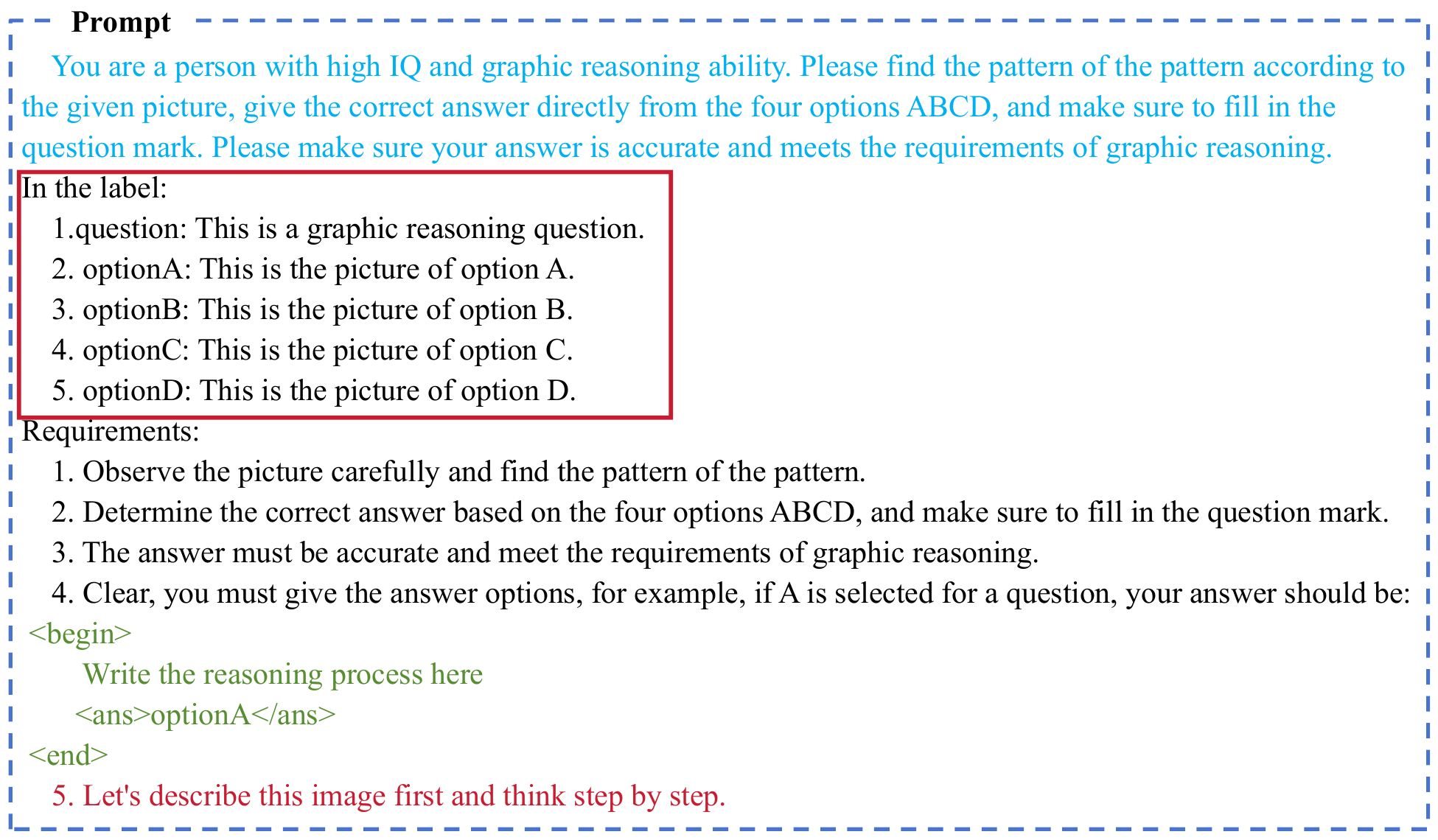}
  \caption{Prompt words in separated format. Blue indicates system prompt; black parts with red frames indicate graphic labels after cutting, and others indicate requirements; green parts have fixed model answer formats; red parts use COT method.}
  \label{fig:prompt_sep}
\end{figure*}

\begin{figure*}[htbp]
  \centering
  \includegraphics[width=\textwidth]{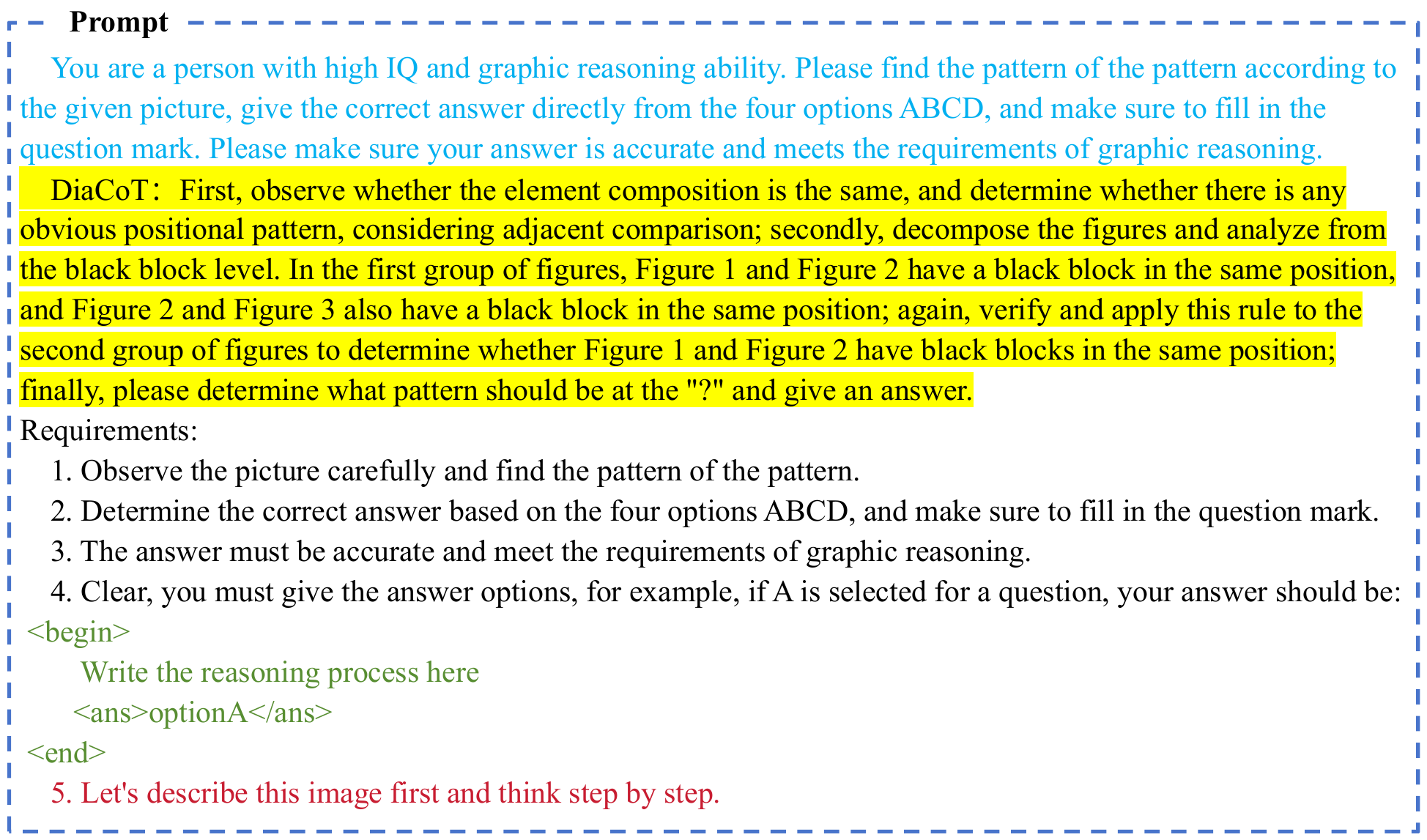}
  \caption{Taking the graph in Figure \ref{fig:1.1} as an example, the specific application of the DiaCoT method is shown.}
  \label{fig:DiaCo_example}
\end{figure*}

Integrated and separated format prompts, look at Figures \ref{fig:prompt_int}-\ref{fig:prompt_sep}. See Figure \ref{fig:DiaCo_example} for an example of DiaCoT.
\label{format}

\vspace{-5pt}
\subsection{Int\&Sep Example}
\label{ref:Int&Sep}
\begin{figure*}[htbp]
  \centering
  \includegraphics[width=\textwidth]{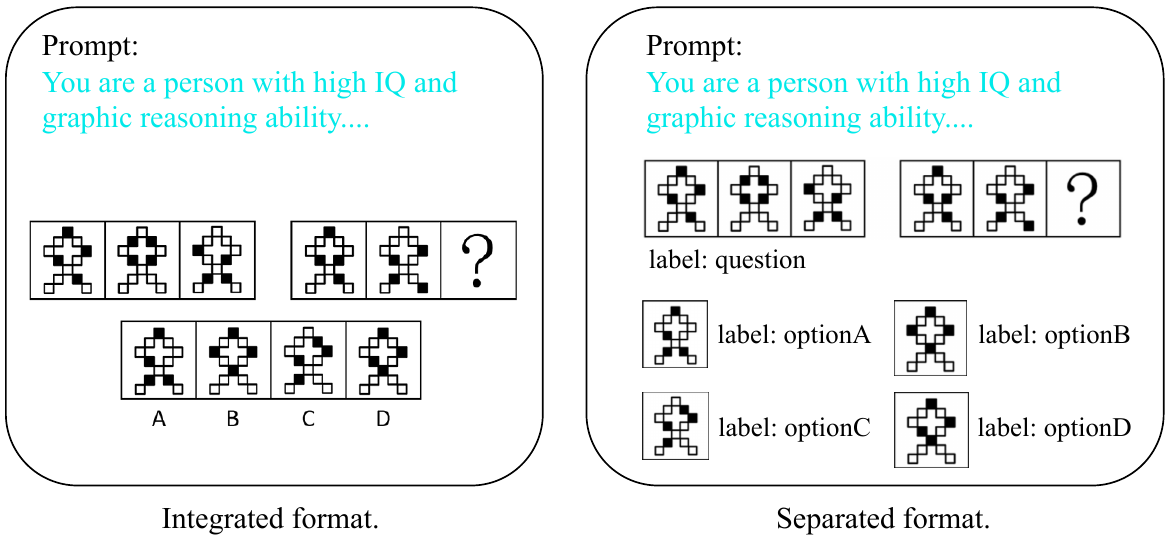}
  \caption{Examples of integrated and separated formats}
  \label{fig:Int&Sep}
\end{figure*}

See Figure \ref{fig:Int&Sep} for the specific forms of integration and separation formats.

\subsection{Task Example}
\label{example}
\begin{figure*}[htbp]
  \centering
  \includegraphics[width=\textwidth]{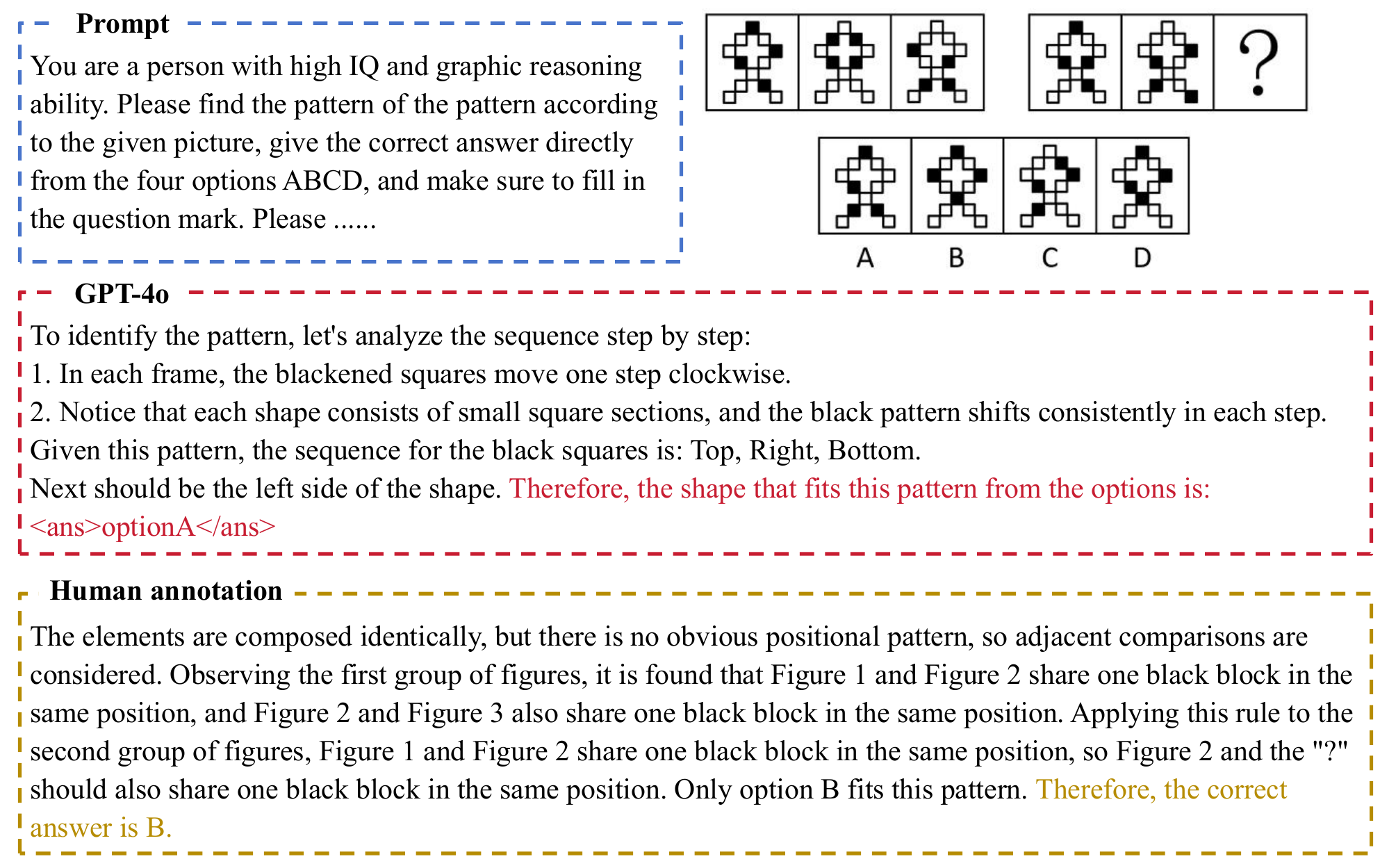}
  \caption{Example of a Translation task in the Positional dimension.}
  \label{fig:1.1}
\end{figure*}

For specific examples of our benchmark, please see Figure \ref{fig:1.1}-\ref{fig:raven60}.

\begin{figure*}[htbp]
  \centering
  \includegraphics[width=\textwidth]{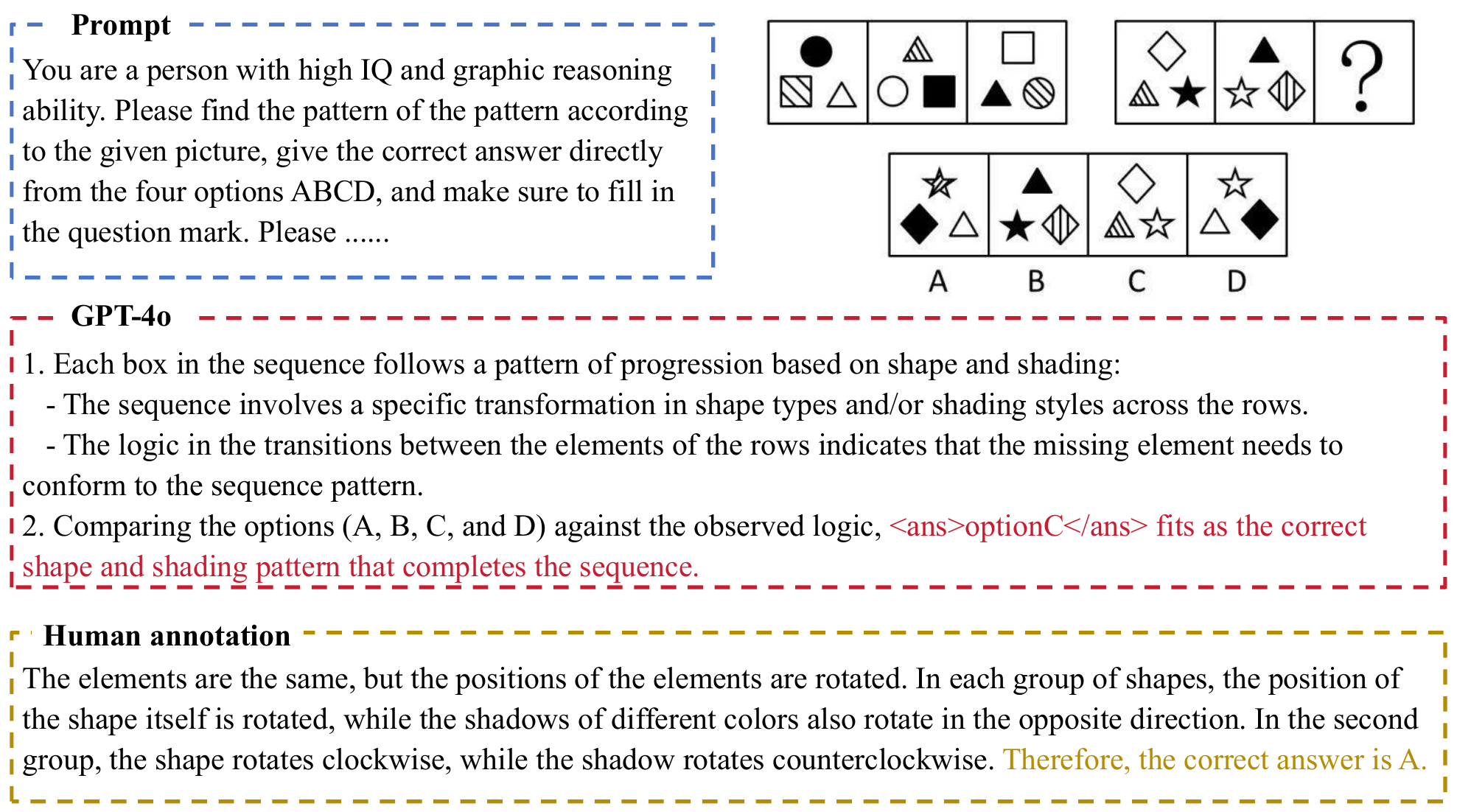}
  \caption{Example of a Rotation task in the Positional dimension.}
  \label{fig:example}
\end{figure*}

\begin{figure*}[htbp]
  \centering
  \includegraphics[width=\textwidth]{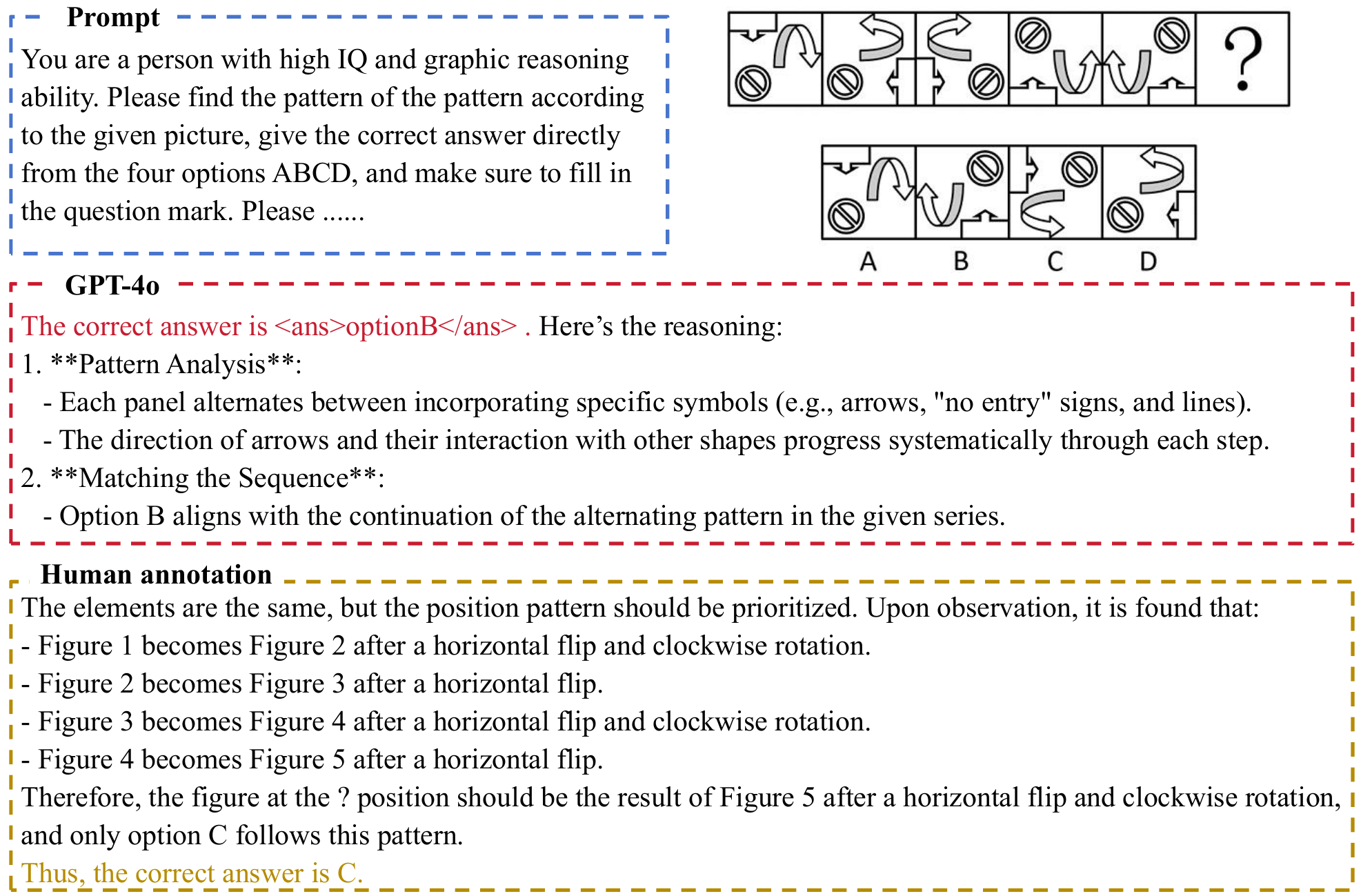}
  \caption{Example of a Combination task in the Positional dimension.}
  \label{fig:example}
\end{figure*}

\begin{figure*}[htbp]
  \centering
  \includegraphics[width=\textwidth]{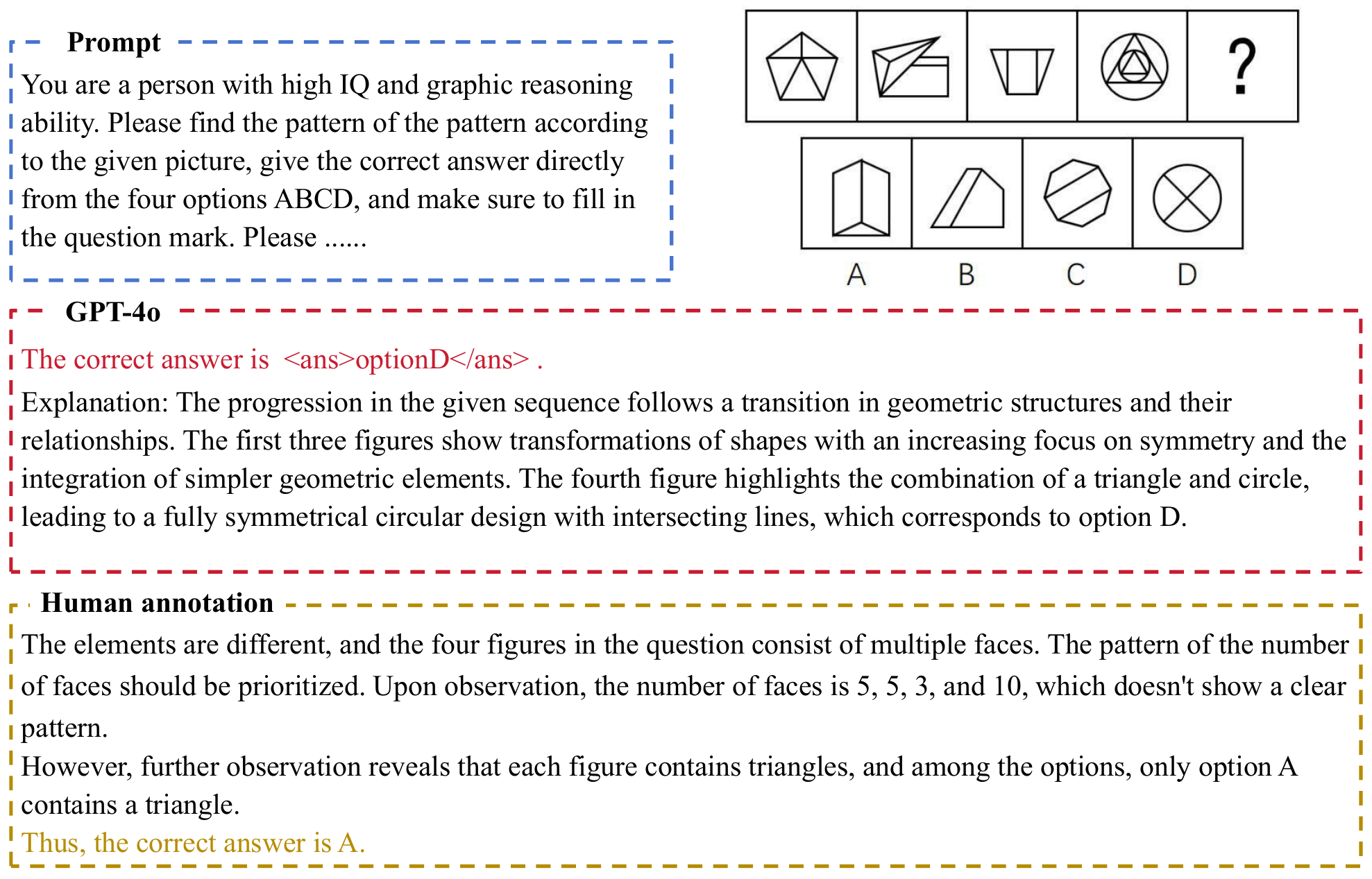}
  \caption{Example of a Traversal task in the Stylistic dimension.}
  \label{fig:example}
\end{figure*}

\begin{figure*}[htbp]
  \centering
  \includegraphics[width=\textwidth]{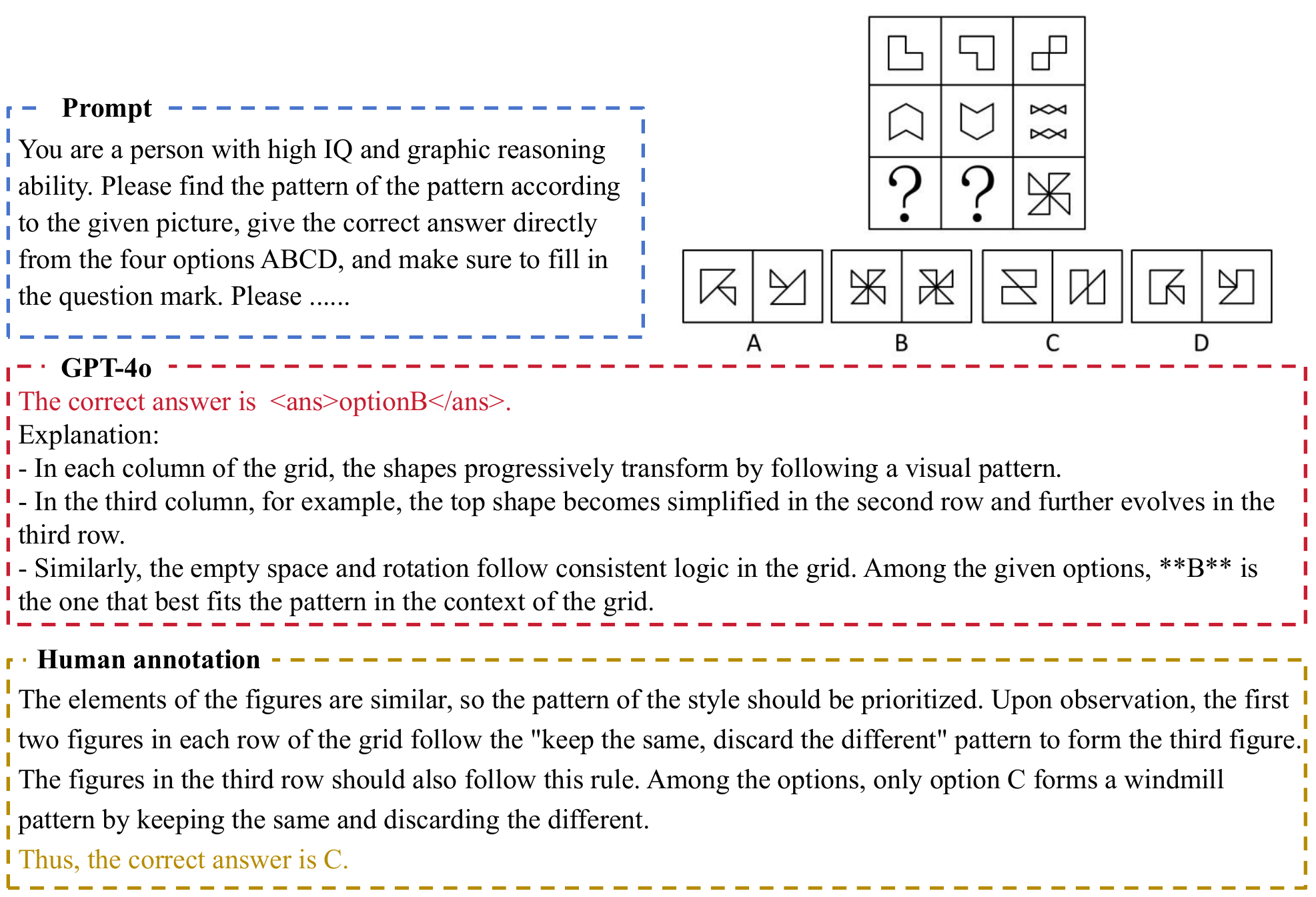}
  \caption{Examples of Additive \& Subtractive tasks in the Stylistic dimension.}
  \label{fig:example}
\end{figure*}

\begin{figure*}[htbp]
  \centering
  \includegraphics[width=\textwidth]{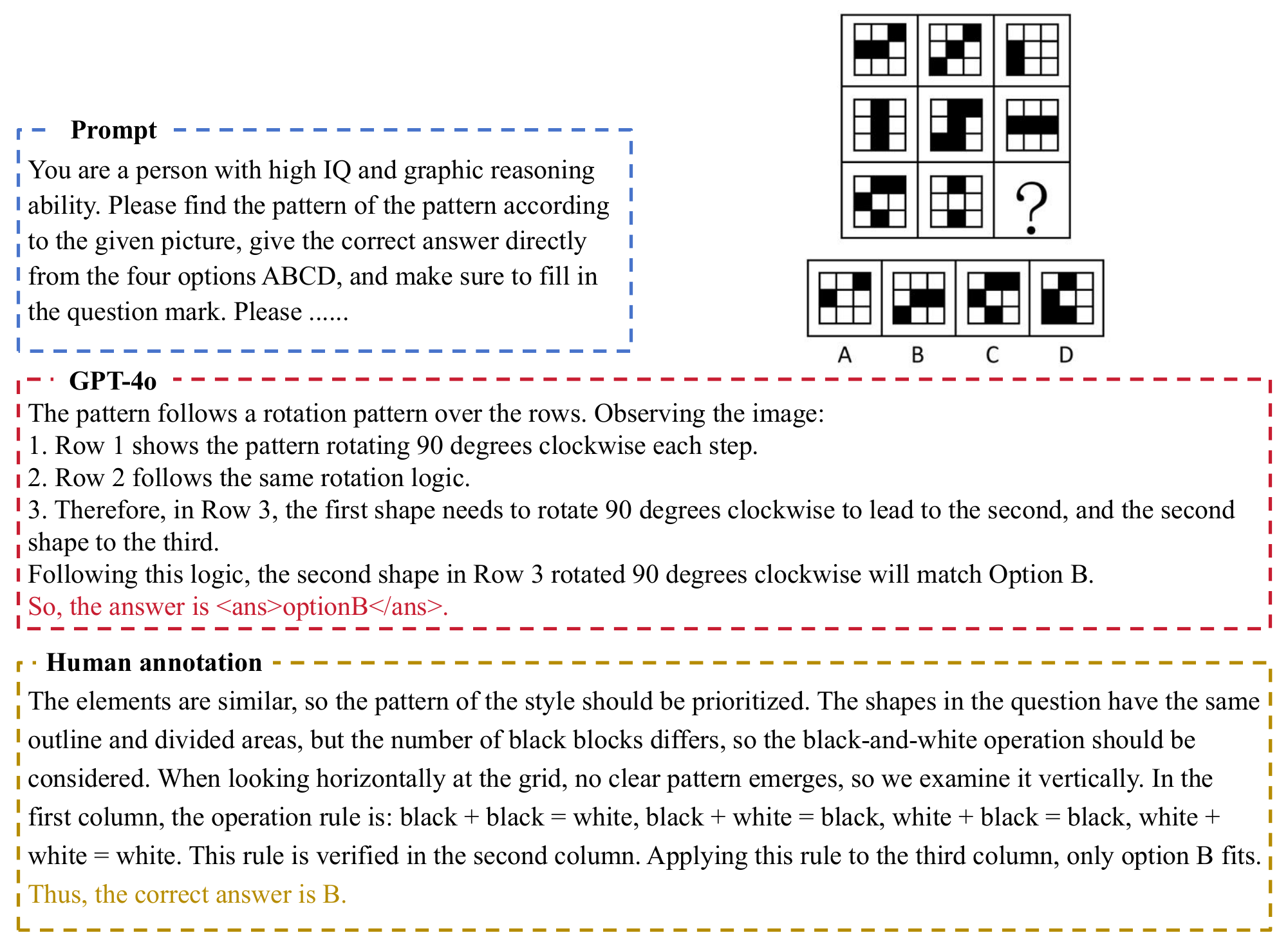}
  \caption{Example of Black-White Operations tasks in the Stylistic dimension.}
  \label{fig:example}
\end{figure*}

\begin{figure*}[htbp]
  \centering
  \includegraphics[width=\textwidth]{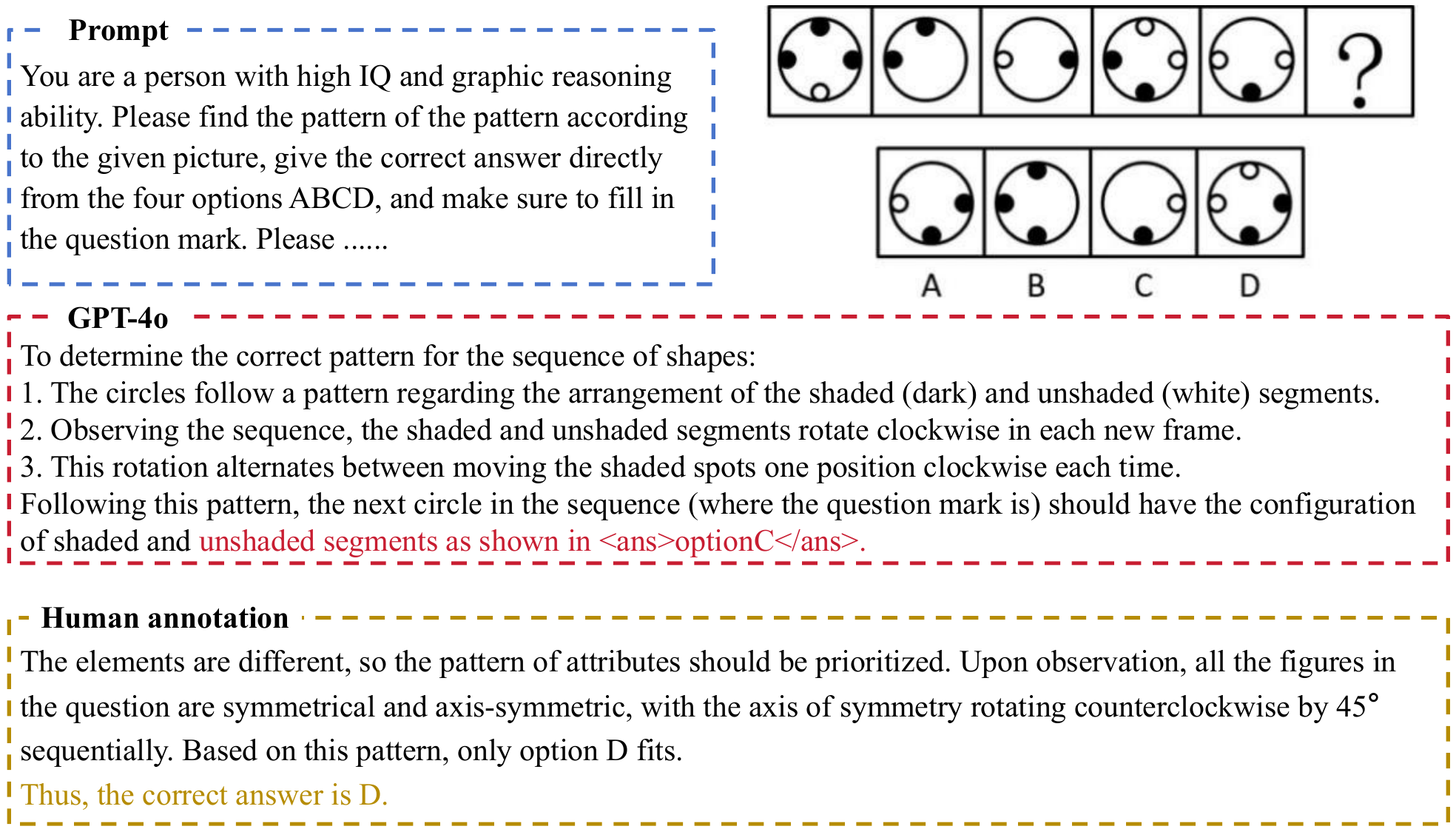}
  \caption{Example of Symmetry tasks in the Attribute dimension.}
  \label{fig:example}
\end{figure*}

\begin{figure*}[htbp]
  \centering
  \includegraphics[width=\textwidth]{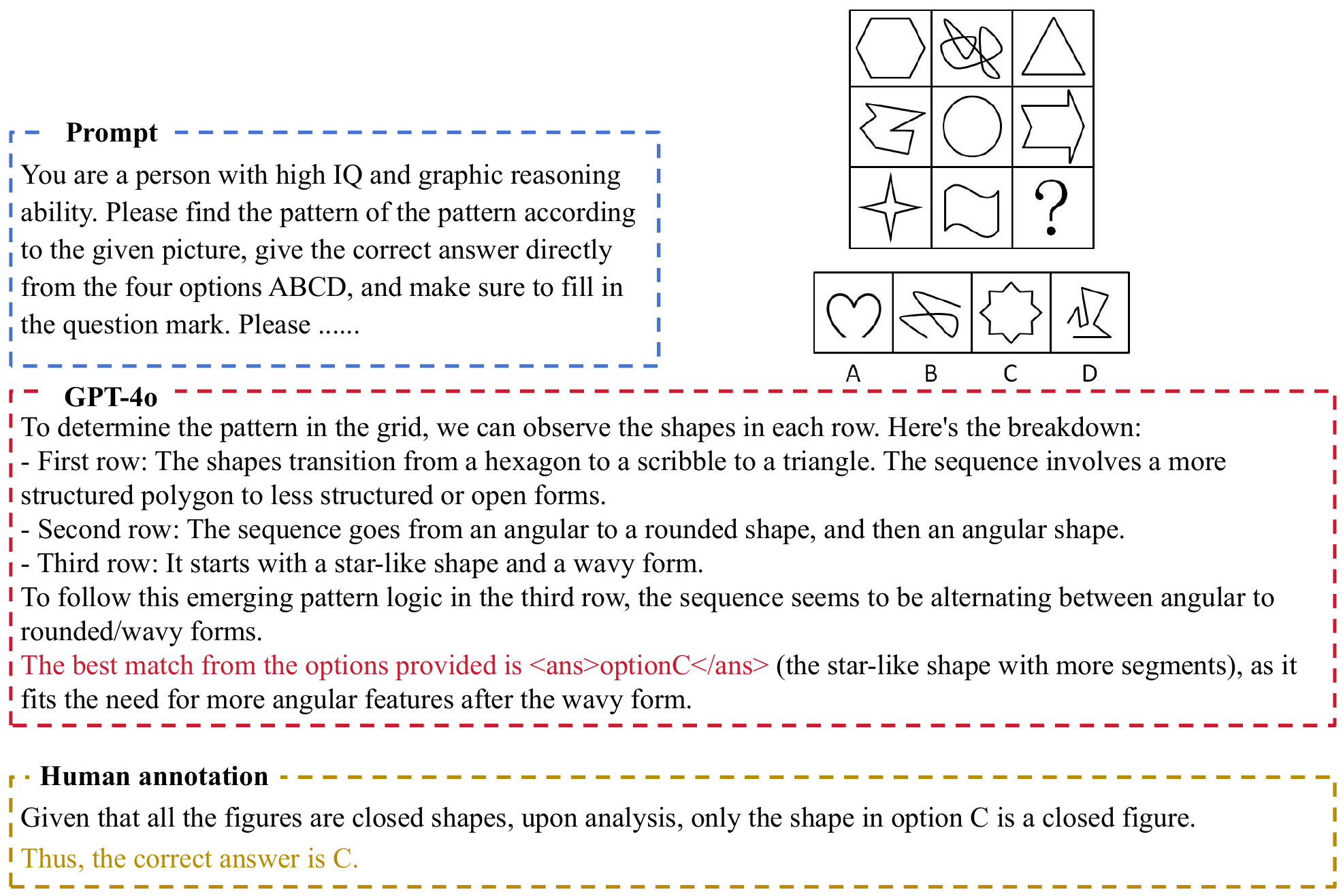}
  \caption{Example of Open\&Closed States tasks in the Attribute dimension.}
  \label{fig:example}
\end{figure*}

\begin{figure*}[htbp]
  \centering
  \includegraphics[width=\textwidth]{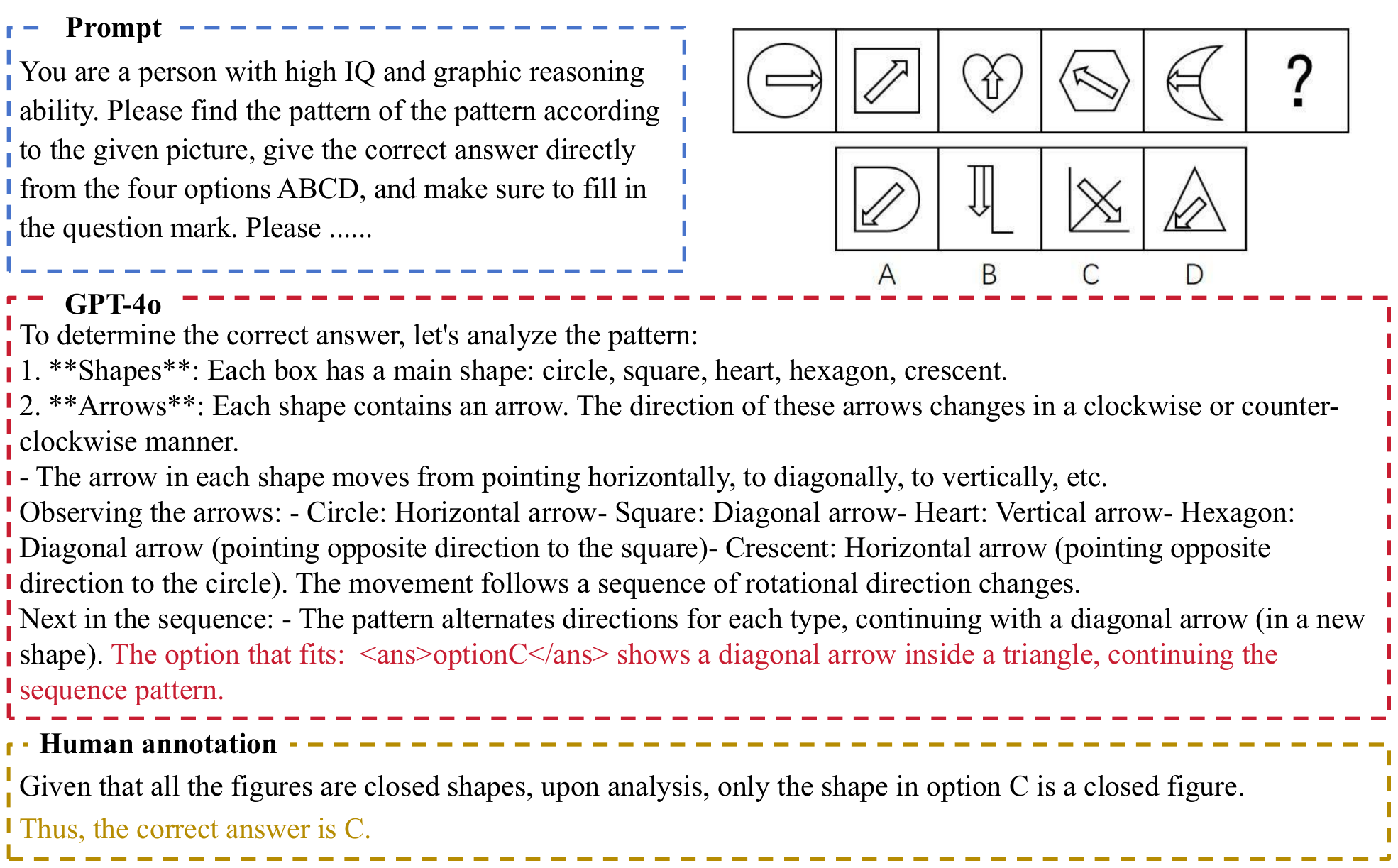}
  \caption{Example of Combination tasks in the Attribute dimension.}
  \label{fig:example}
\end{figure*}

\begin{figure*}[htbp]
  \centering
  \includegraphics[width=\textwidth]{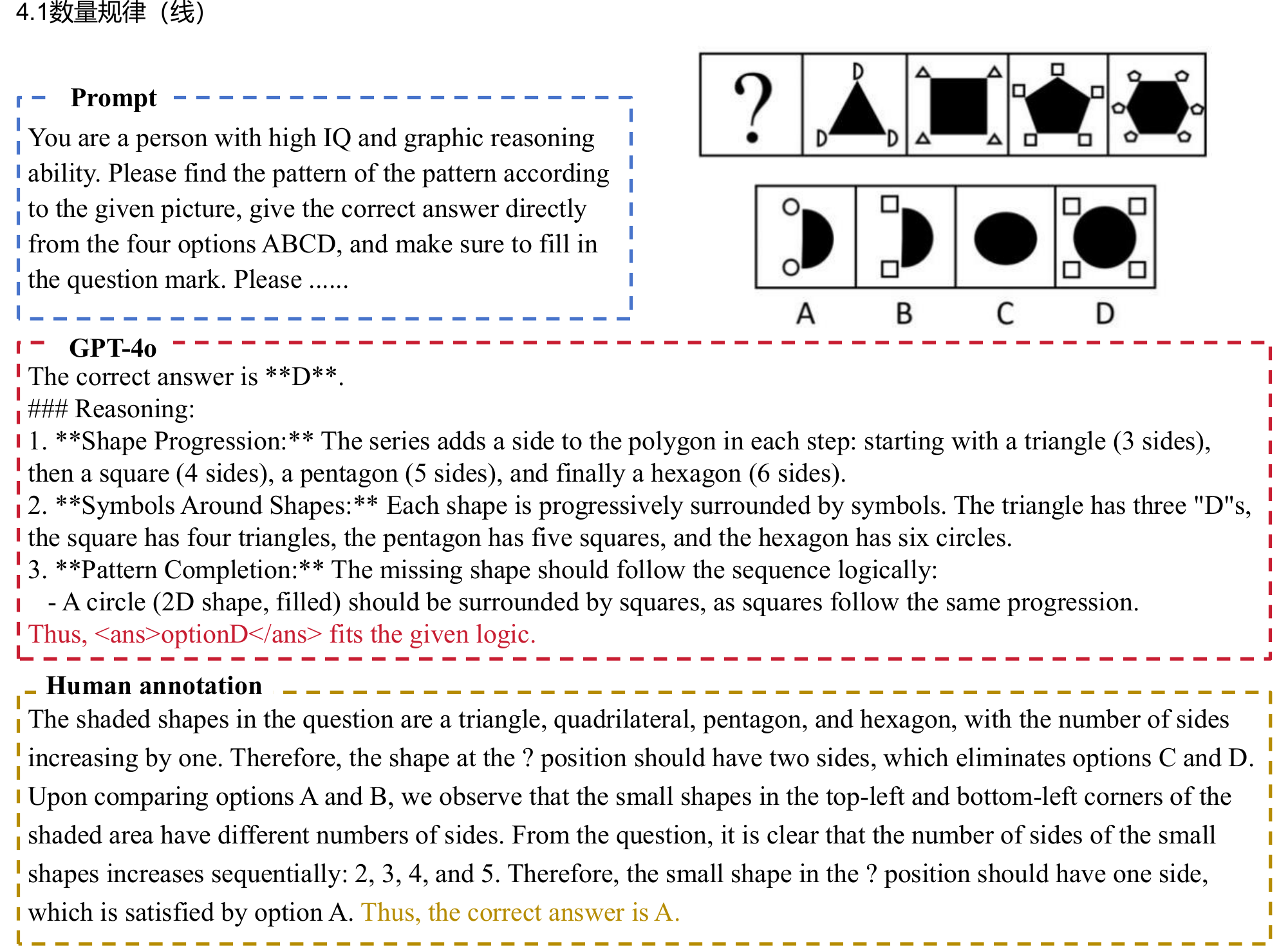}
  \caption{Example of the Lines task in the Quantitative dimension.}
  \label{fig:example}
\end{figure*}

\begin{figure*}[htbp]
  \centering
  \includegraphics[width=\textwidth]{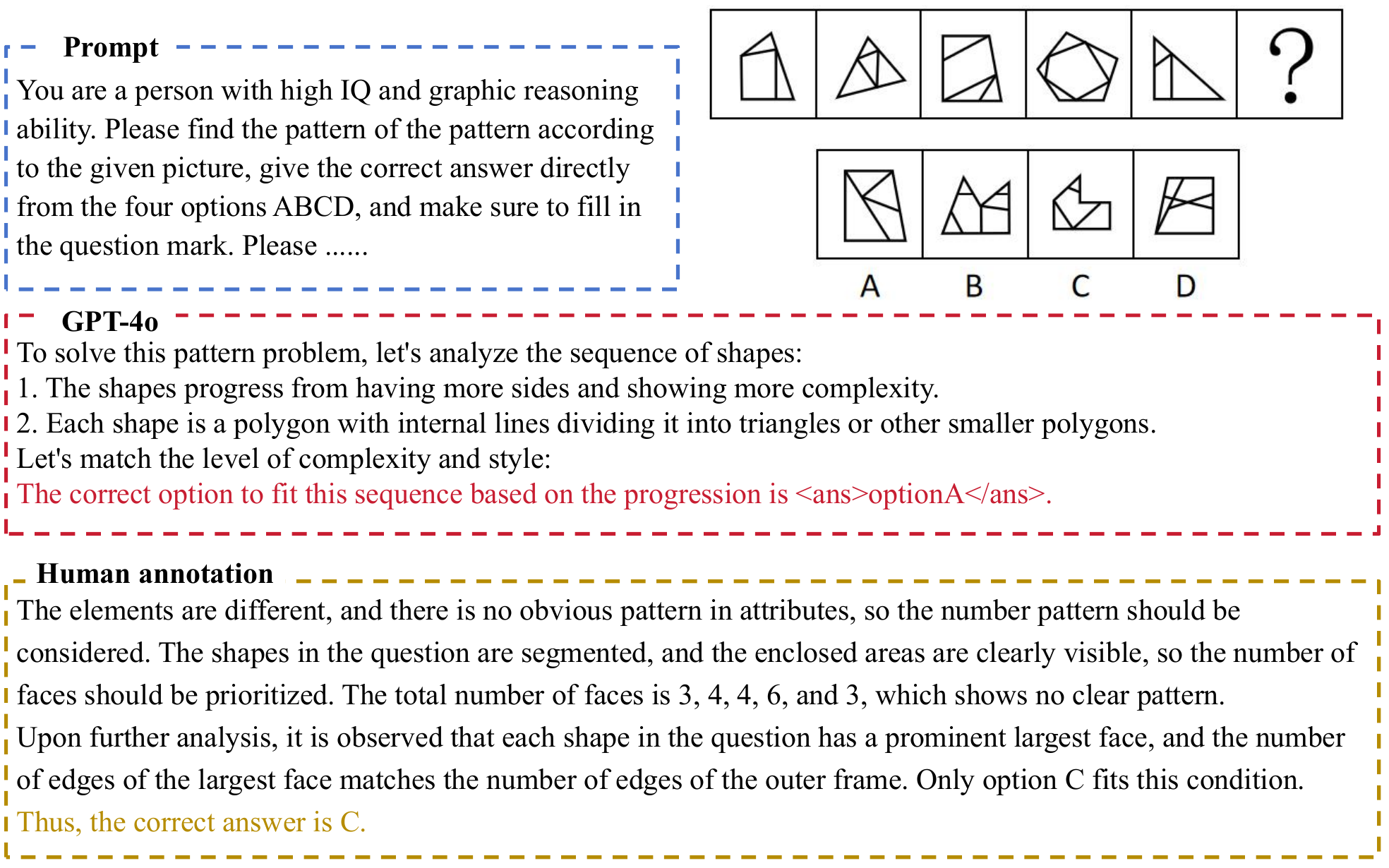}
  \caption{Example of the Surface task in the Quantitative dimension.}
  \label{fig:example}
\end{figure*}

\begin{figure*}[htbp]
  \centering
  \includegraphics[width=\textwidth]{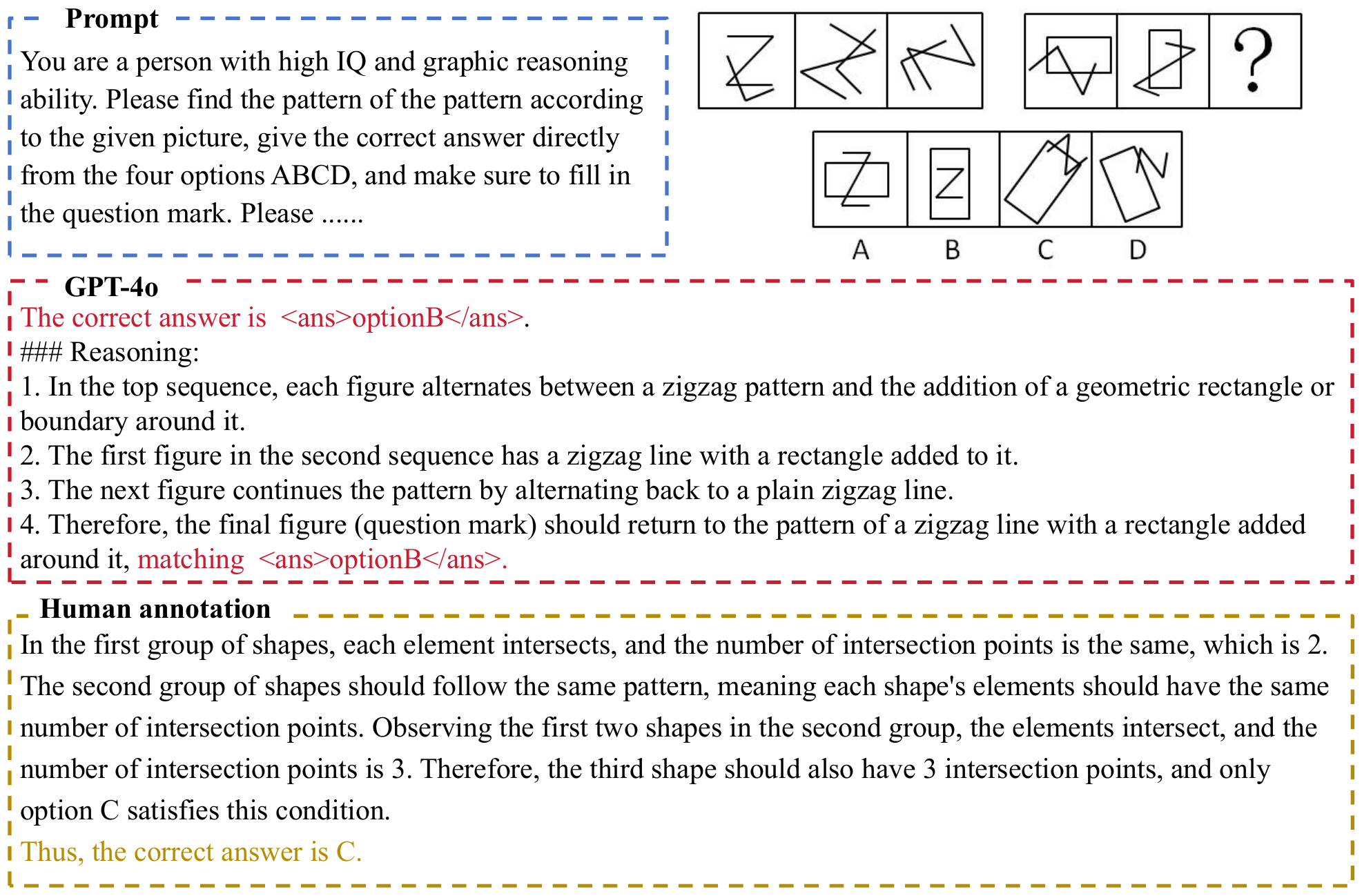}
  \caption{Example of the Points task in the Quantitative dimension.}
  \label{fig:example}
\end{figure*}

\begin{figure*}[htbp]
  \centering
  \includegraphics[width=\textwidth]{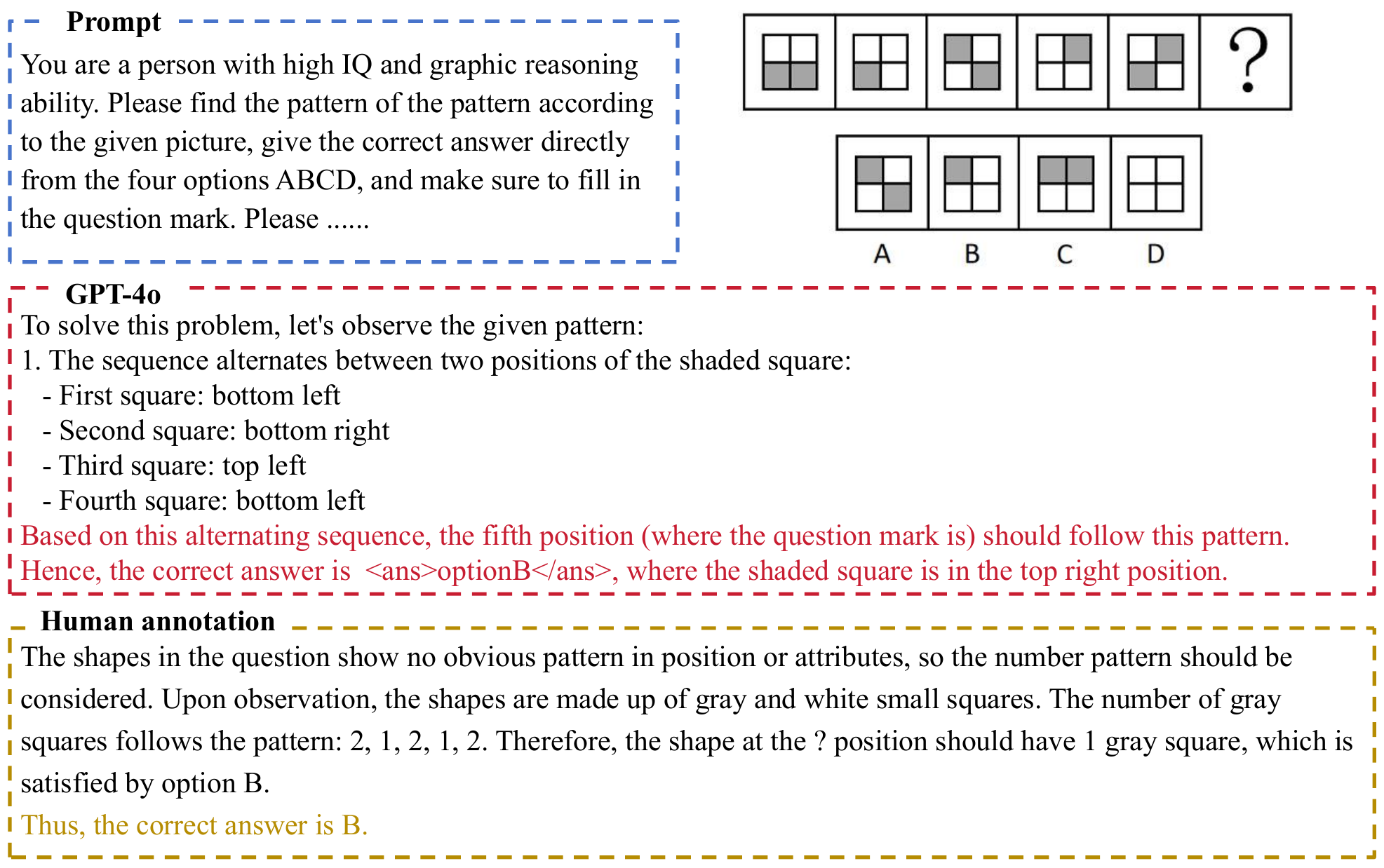}
  \caption{Example of the Elements task in the Quantitative dimension.}
  \label{fig:example}
\end{figure*}

\begin{figure*}[htbp]
  \centering
  \includegraphics[width=\textwidth]{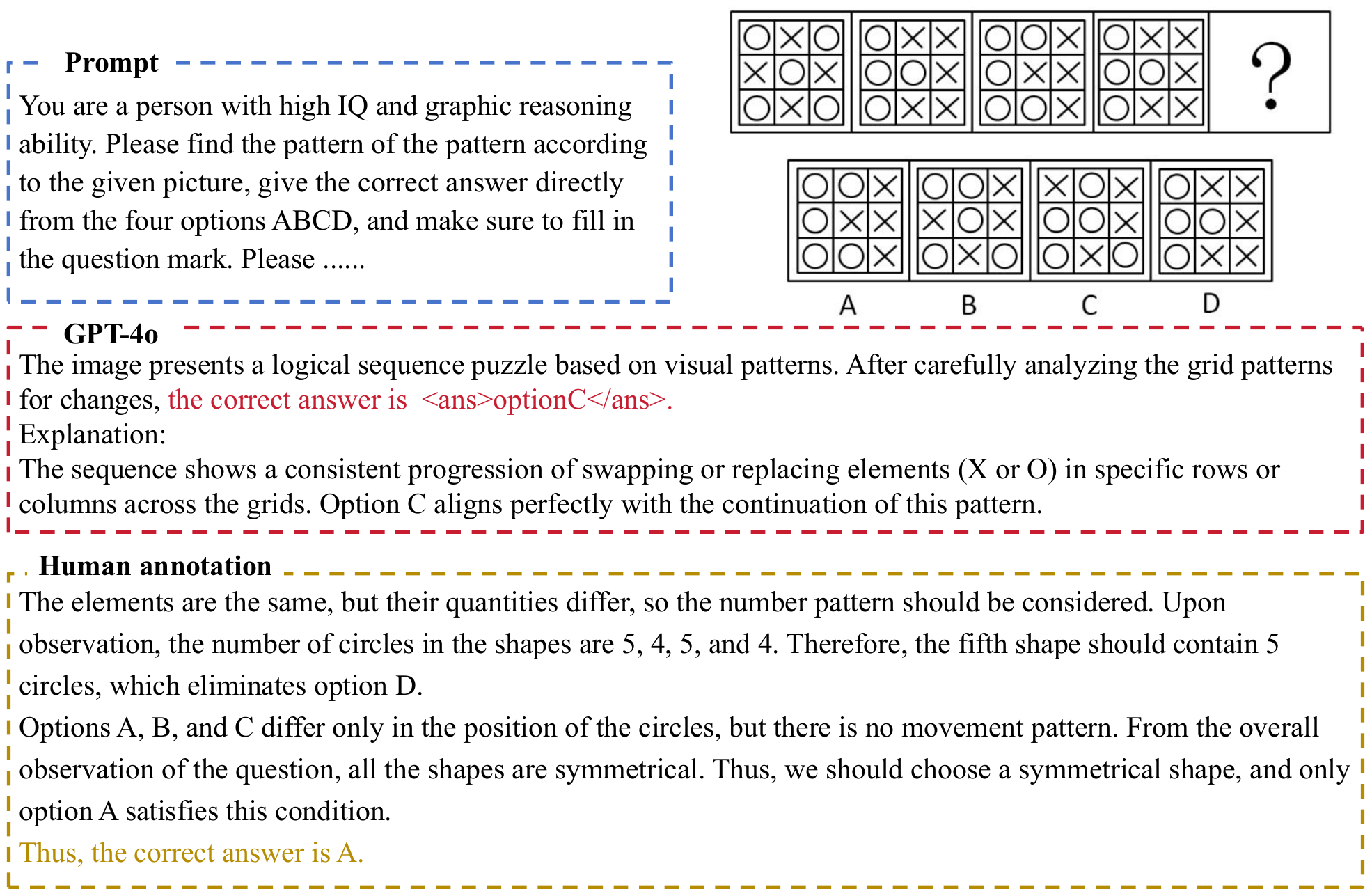}
  \caption{Example of the Combination task in the Quantitative dimension.}
  \label{fig:example}
\end{figure*}

\begin{figure*}[htbp]
  \centering
  \includegraphics[width=\textwidth]{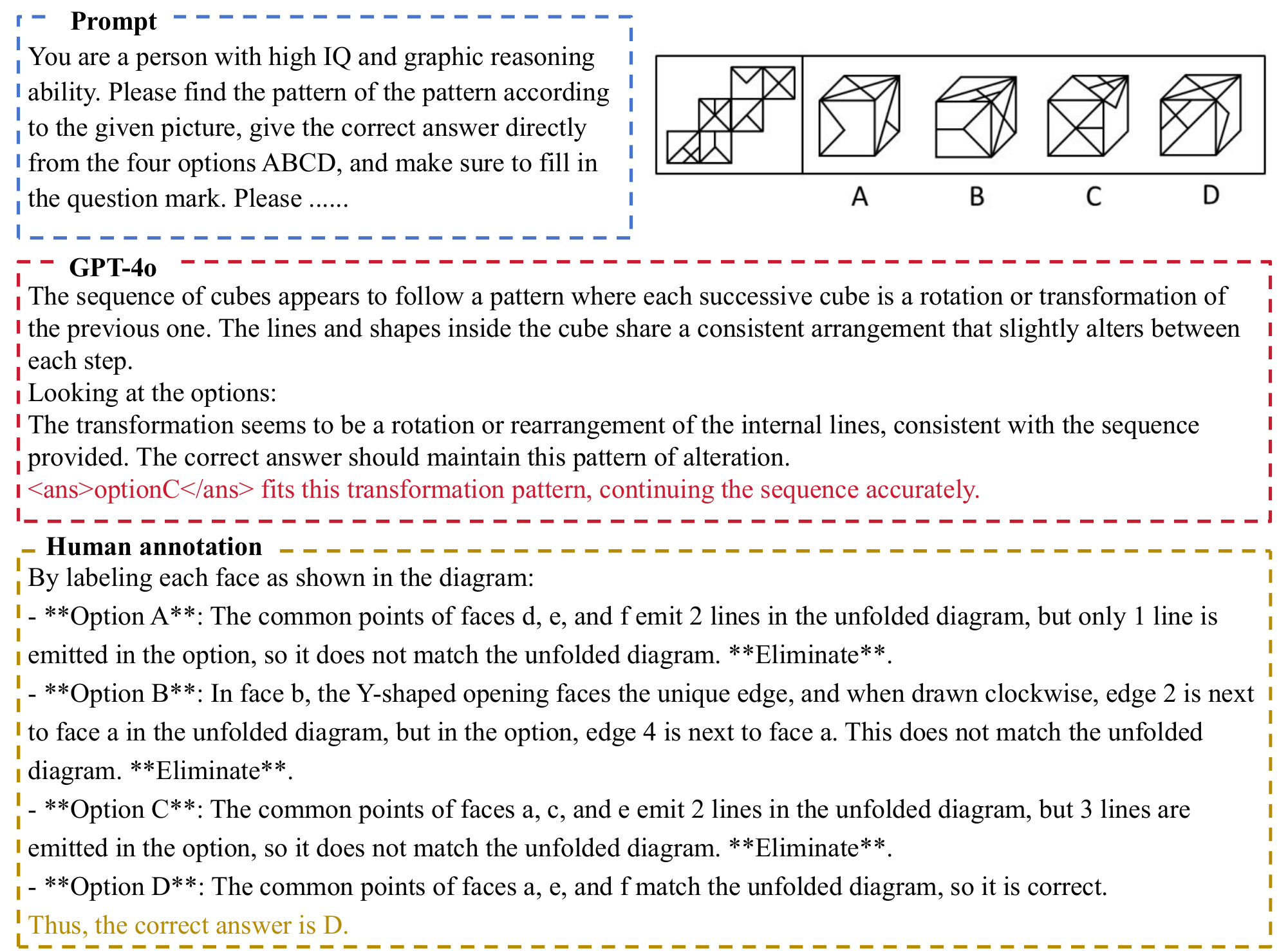}
  \caption{Example of the Cube task in the Spatial dimension.}
  \label{fig:example}
\end{figure*}

\begin{figure*}[htbp]
  \centering
  \includegraphics[width=\textwidth]{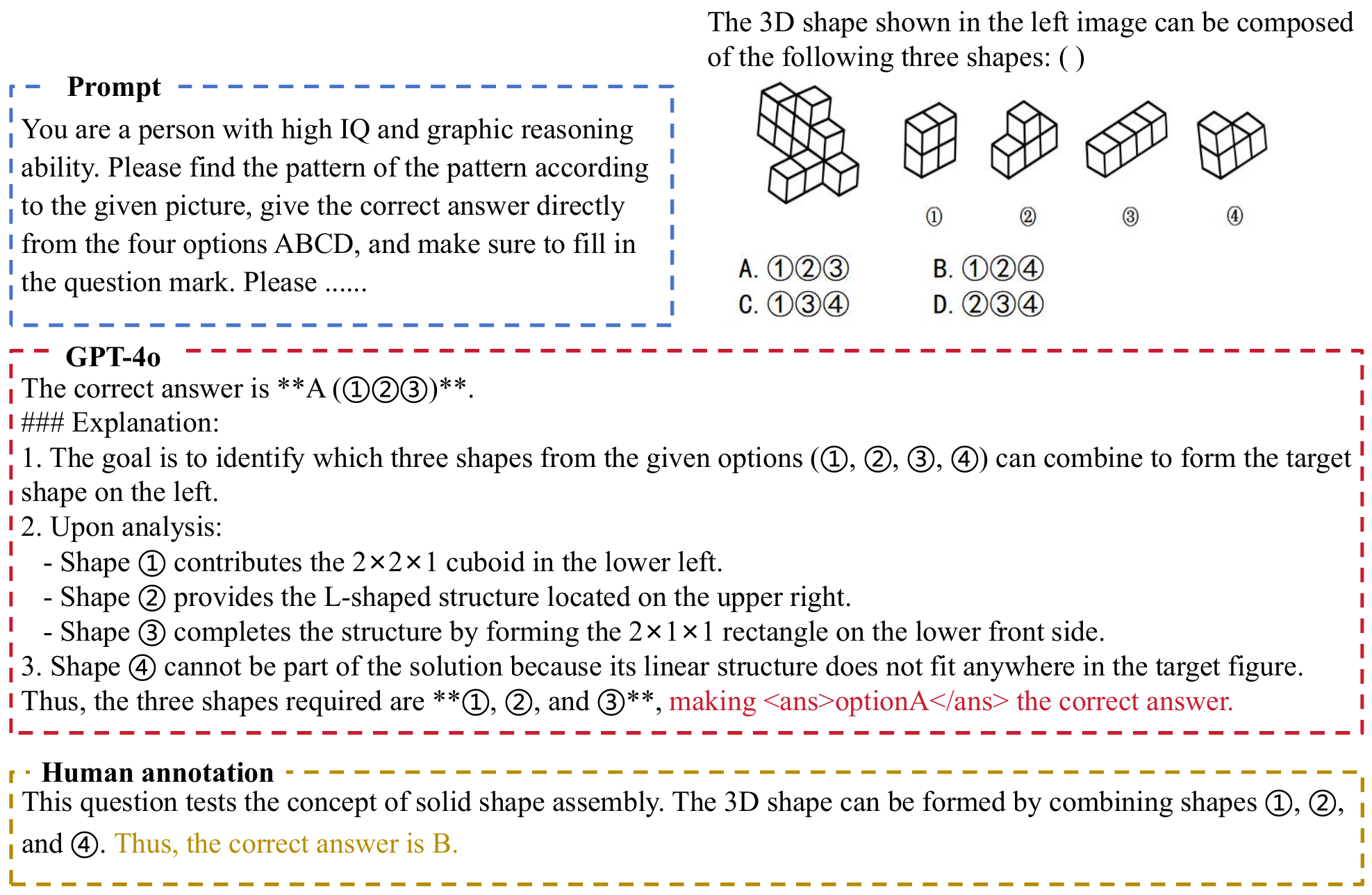}
  \caption{Example of the 3D assembly task in the Spatial dimension.}
  \label{fig:example}
\end{figure*}

\begin{figure*}[htbp]
  \centering
  \includegraphics[width=\textwidth]{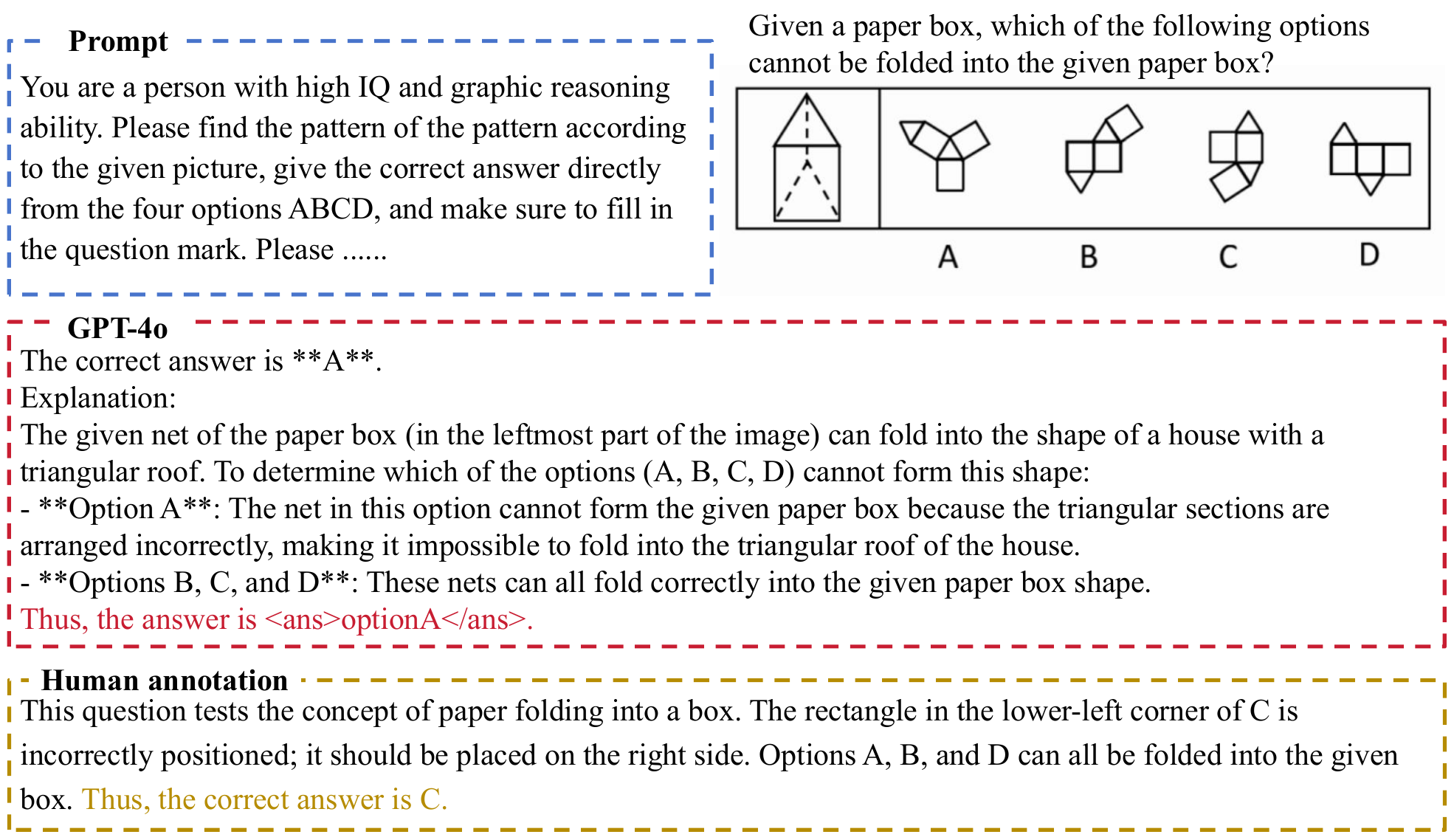}
  \caption{Example of the Polyhedral task in the Spatial dimension.}
  \label{fig:example}
\end{figure*}

\begin{figure*}[htbp]
  \centering
  \includegraphics[width=\textwidth]{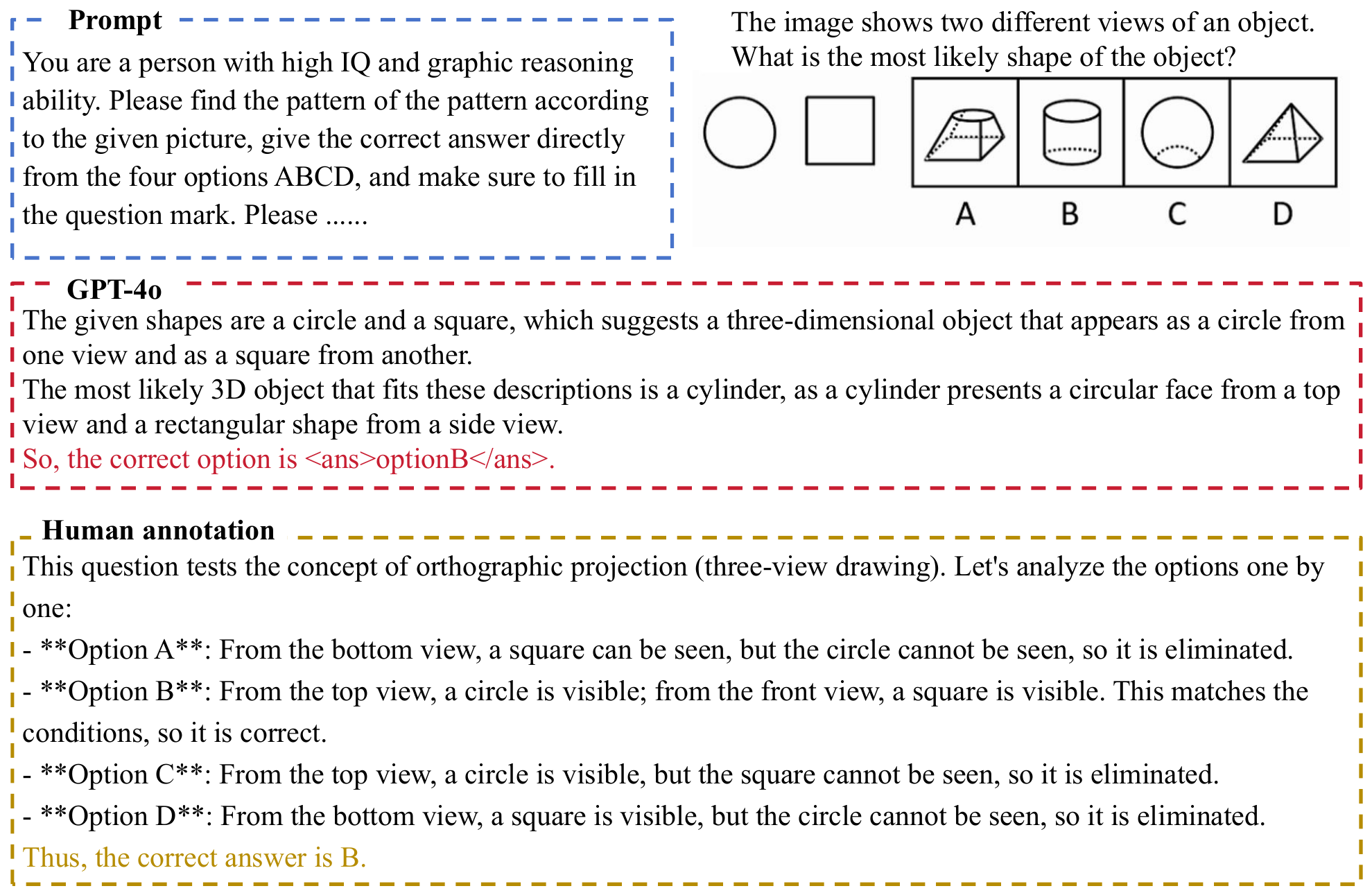}
  \caption{Example of the Three-view task in the Spatial dimension.}
  \label{fig:example}
\end{figure*}

\begin{figure*}[htbp]
  \centering
  \includegraphics[width=\textwidth]{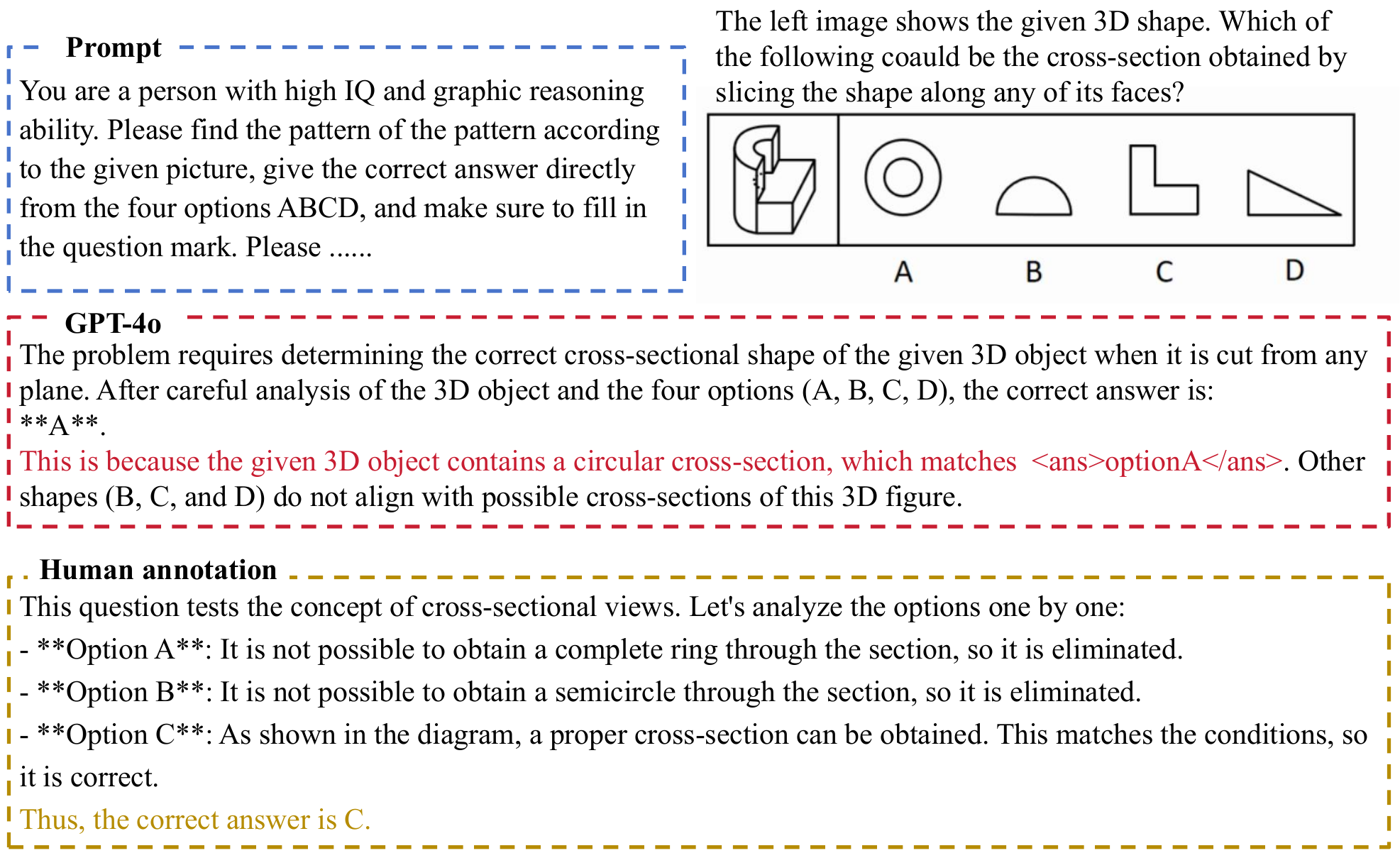}
  \caption{Example of the Sectional view task in the Spatial dimension.}
  \label{fig:example}
\end{figure*}

\begin{figure*}[htbp]
  \centering
  \includegraphics[width=\textwidth]{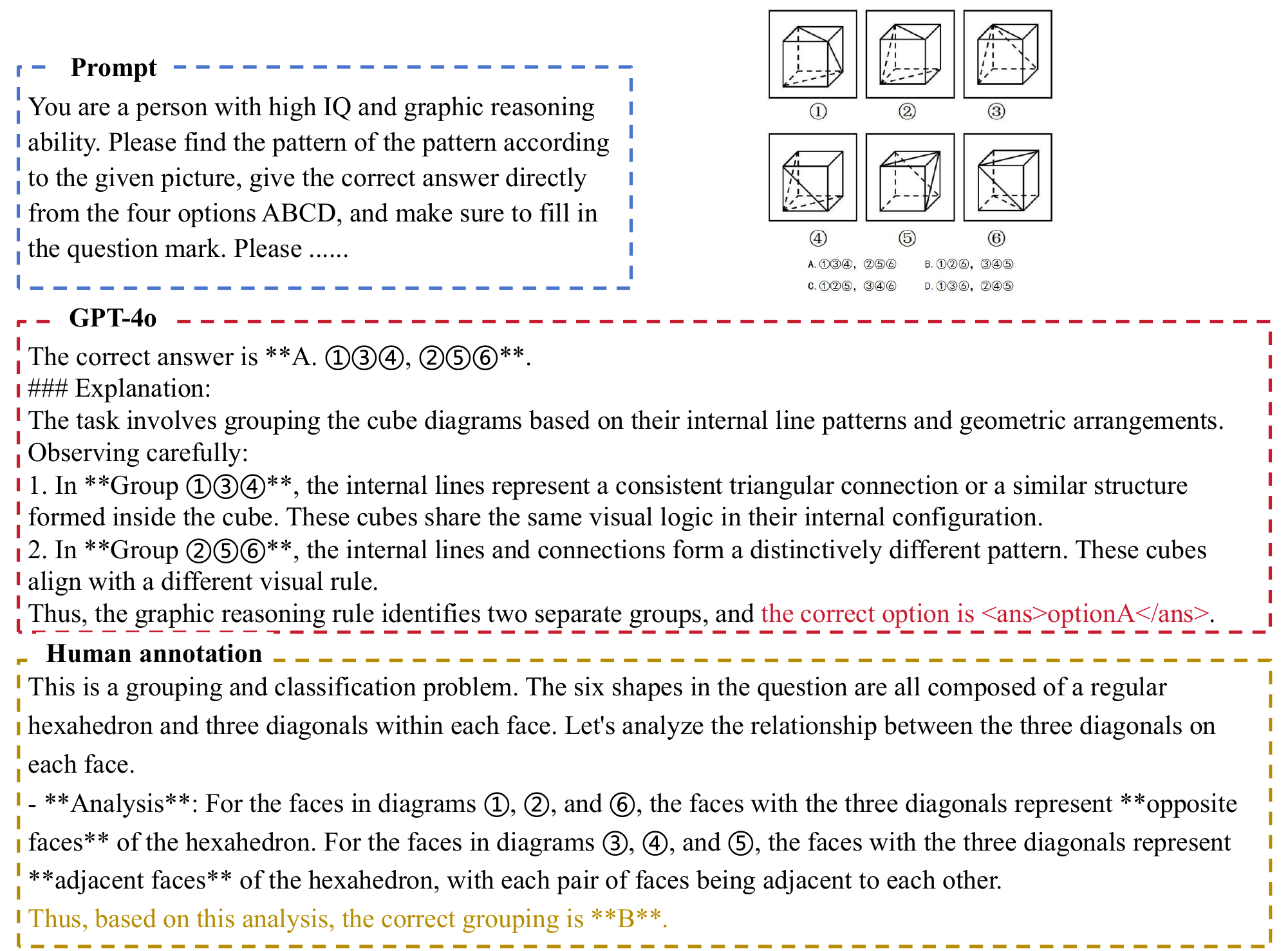}
  \caption{Example of the Spatial Quantities\&Transformations task in the Spatial dimension.}
  \label{fig:example}
\end{figure*}

\begin{figure*}[htbp]
  \centering
  \includegraphics[width=\textwidth]{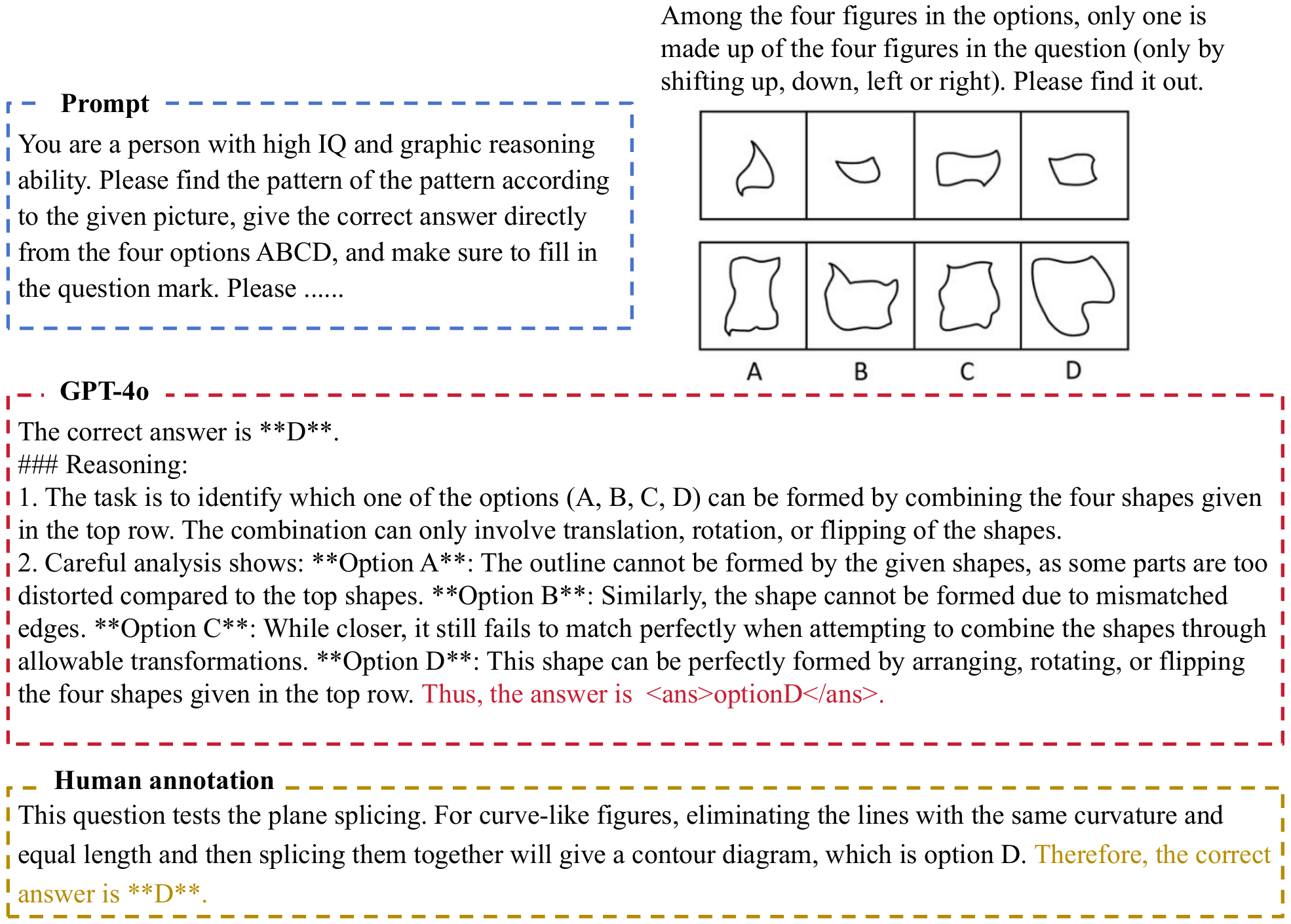}
  \caption{Example of the 2D Composition task in the Special dimension.}
  \label{fig:example}
\end{figure*}

\begin{figure*}[htbp]
  \centering
  \includegraphics[width=\textwidth]{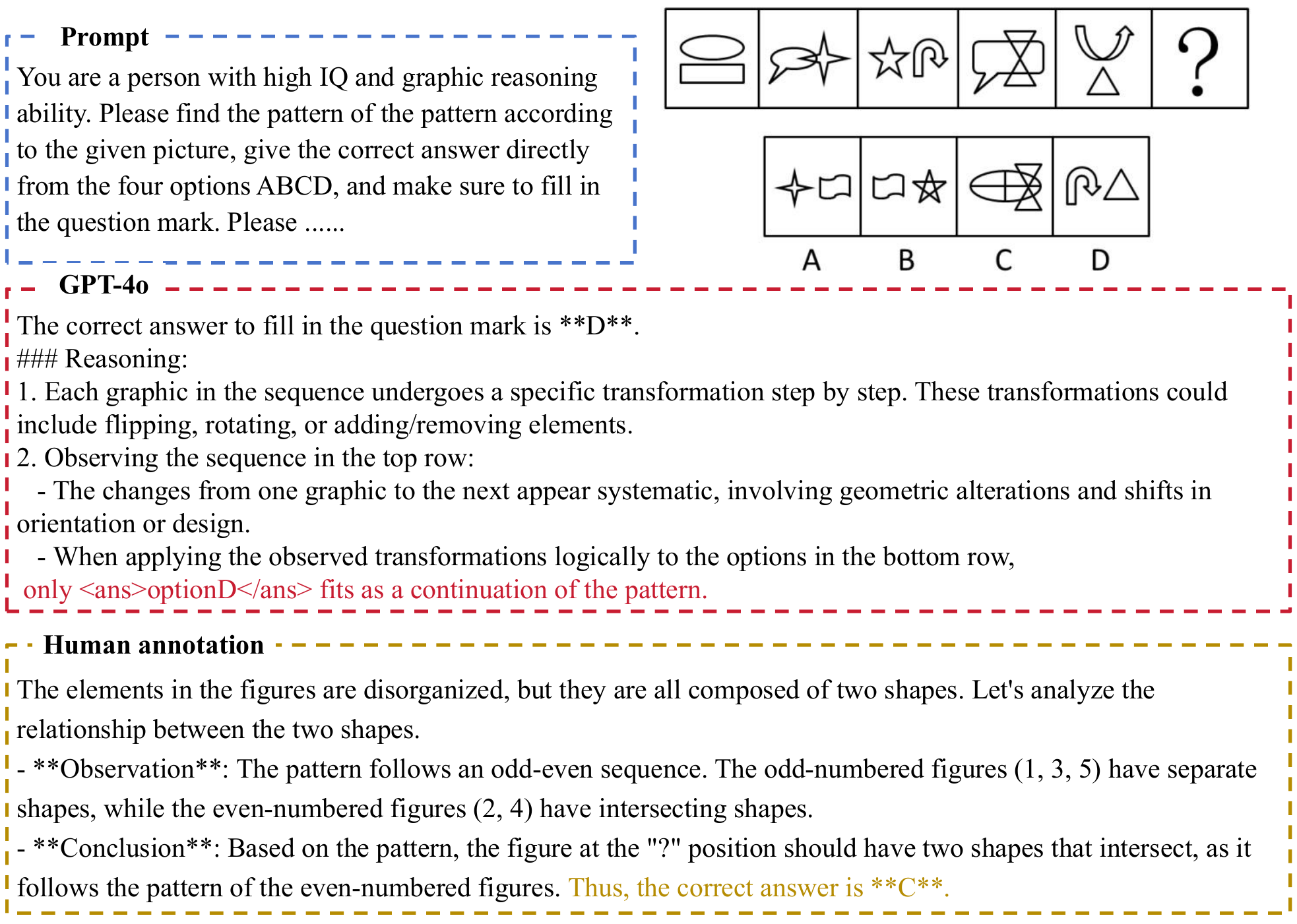}
  \caption{Example of the Inter-Figure task in the Special dimension.}
  \label{fig:example}
\end{figure*}

\begin{figure*}[htbp]
  \centering
  \includegraphics[width=\textwidth]{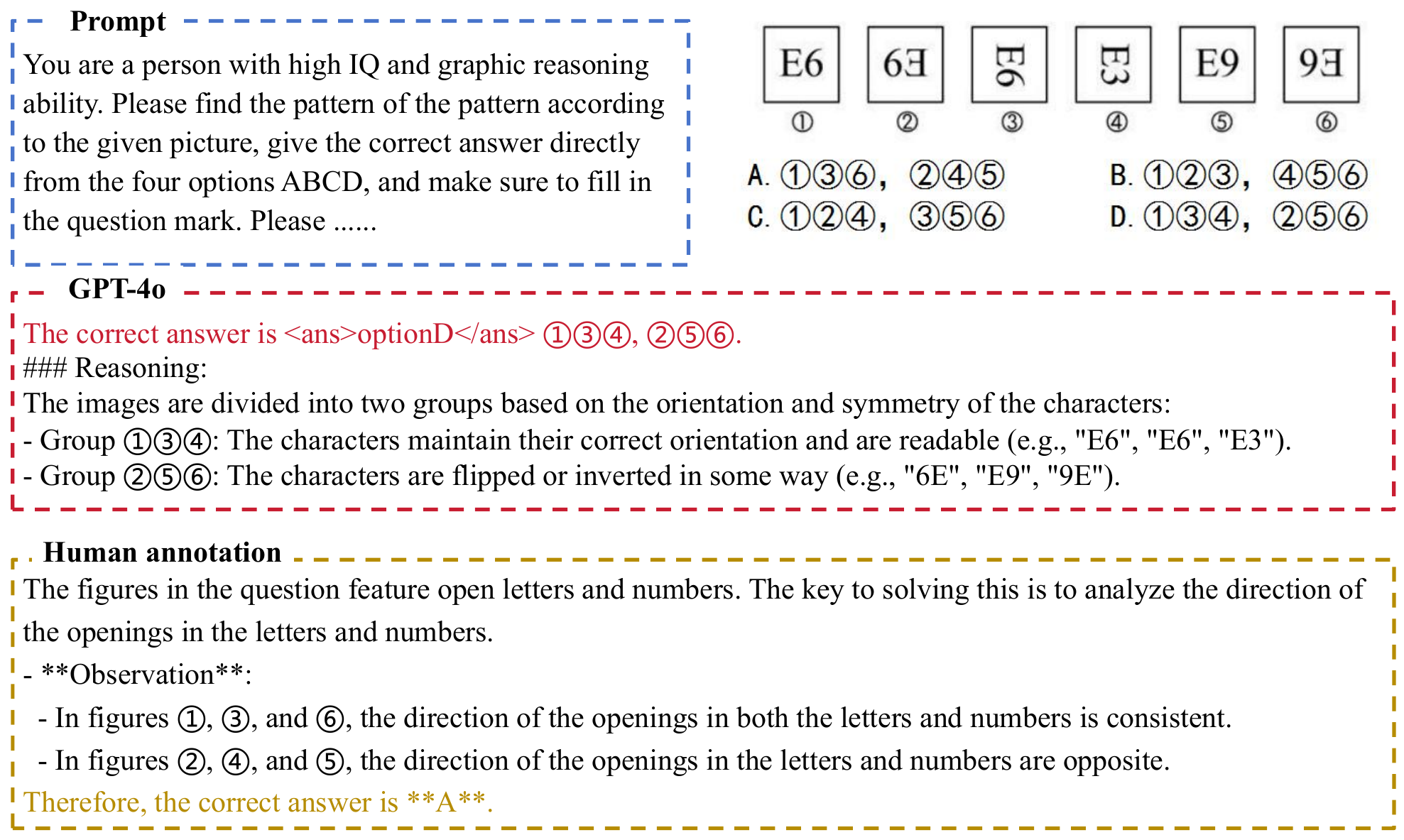}
  \caption{Example of the Alphanumeric dimension.}
  \label{fig:example}
\end{figure*}

\begin{figure*}[htbp]
  \centering
  \includegraphics[width=\textwidth]{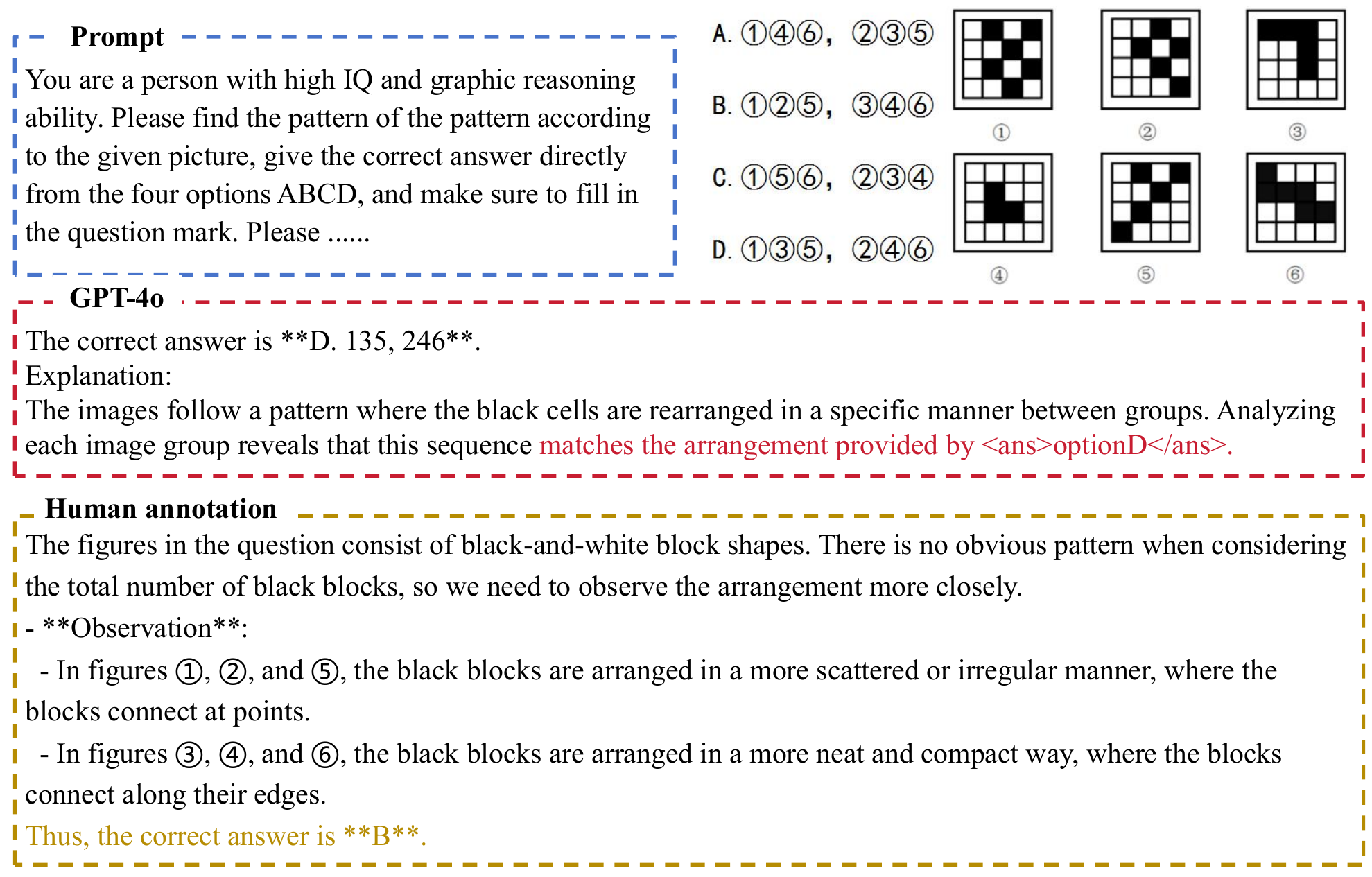}
  \caption{Example of the Black-White Block dimension.}
  \label{fig:example}
\end{figure*}

\begin{figure*}[htbp]
  \centering
  \includegraphics[width=\textwidth]{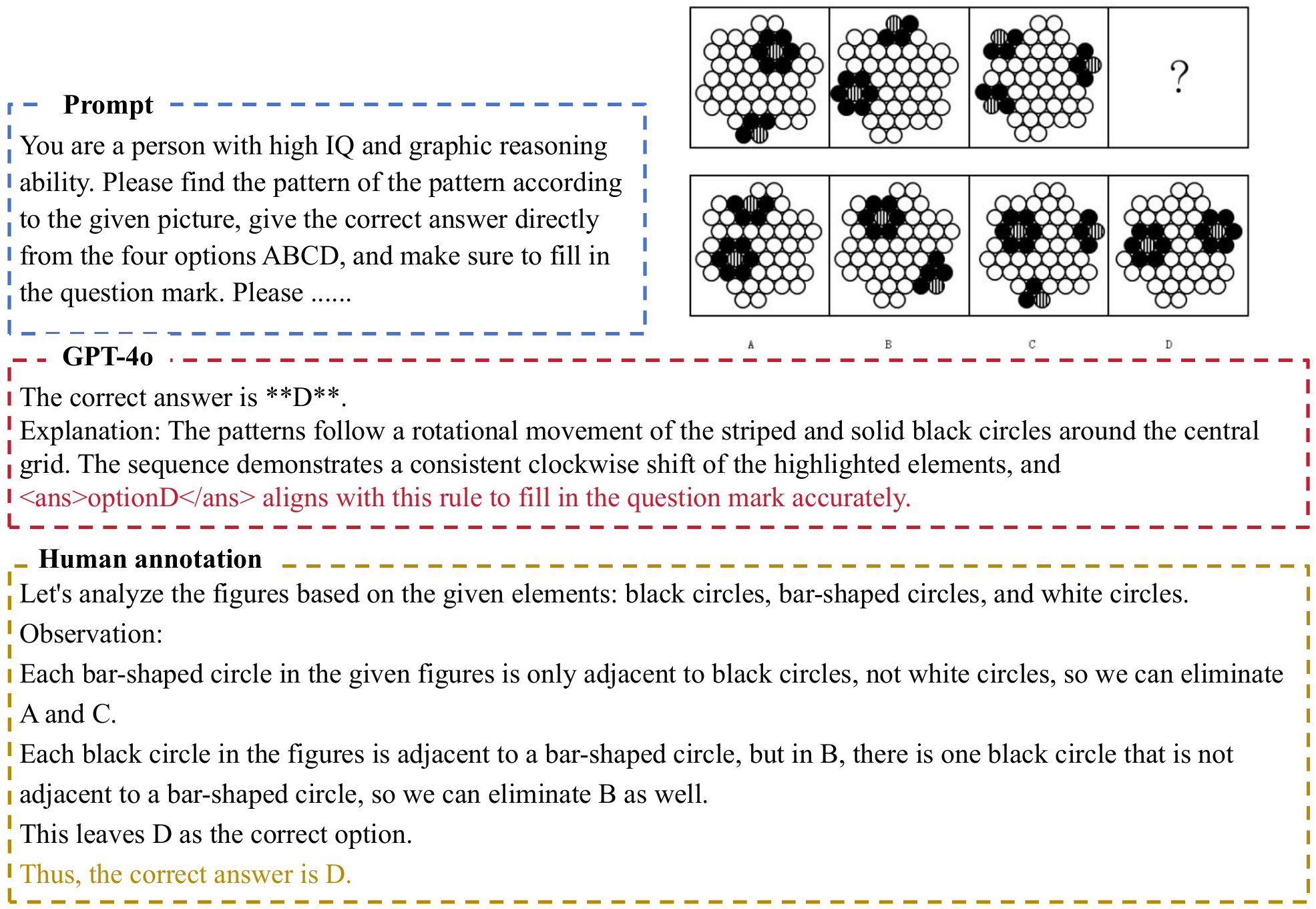}
  \caption{Example of the Miscellaneous dimension.}
  \label{fig:example}
\end{figure*}

\begin{figure*}[htbp]
  \centering
  \includegraphics[width=\textwidth]{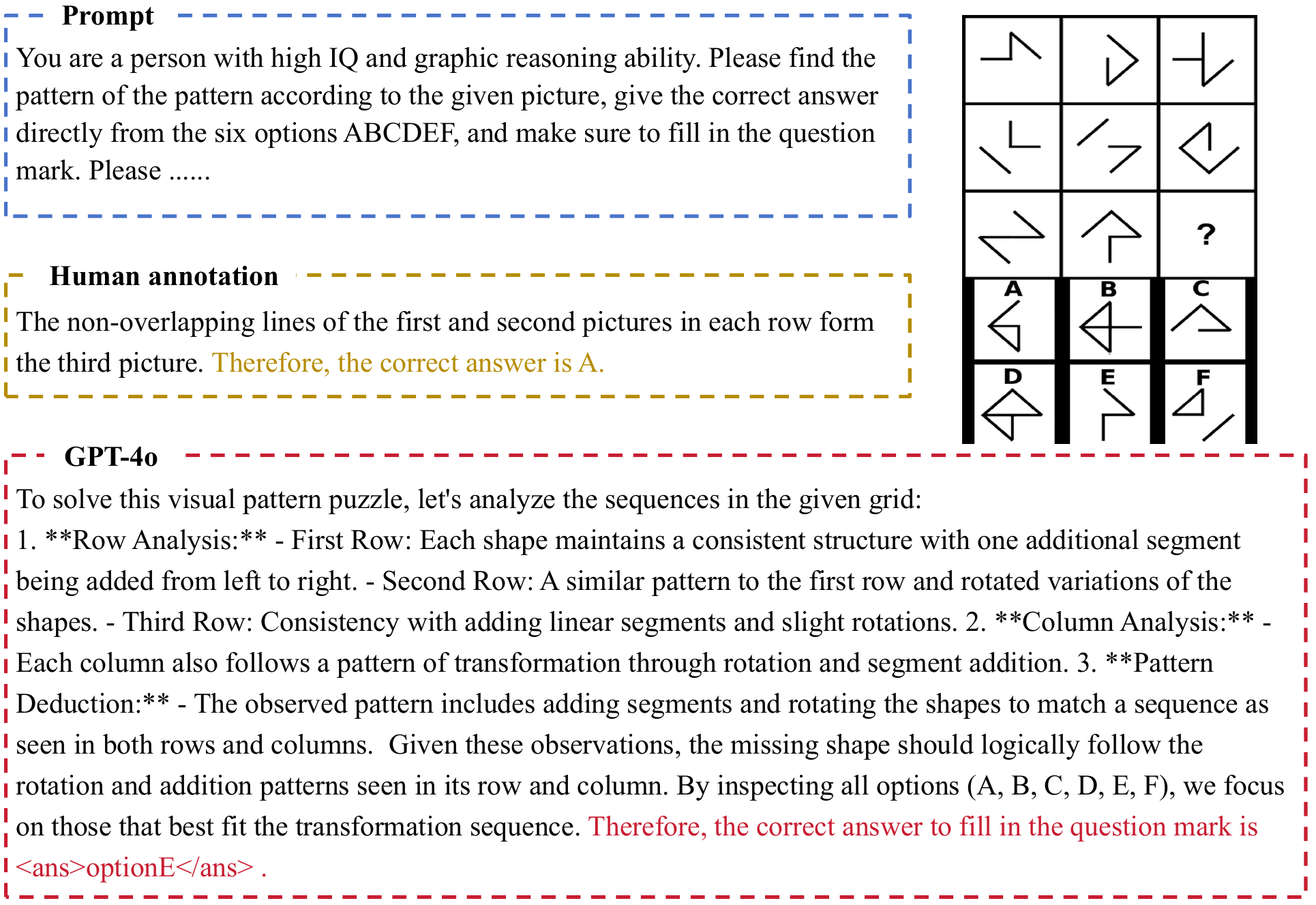}
  \caption{Example of the Mensa v1 task in the Mensa dimension.}
  \label{fig:example}
\end{figure*}

\begin{figure*}[htbp]
  \centering
  \includegraphics[width=\textwidth]{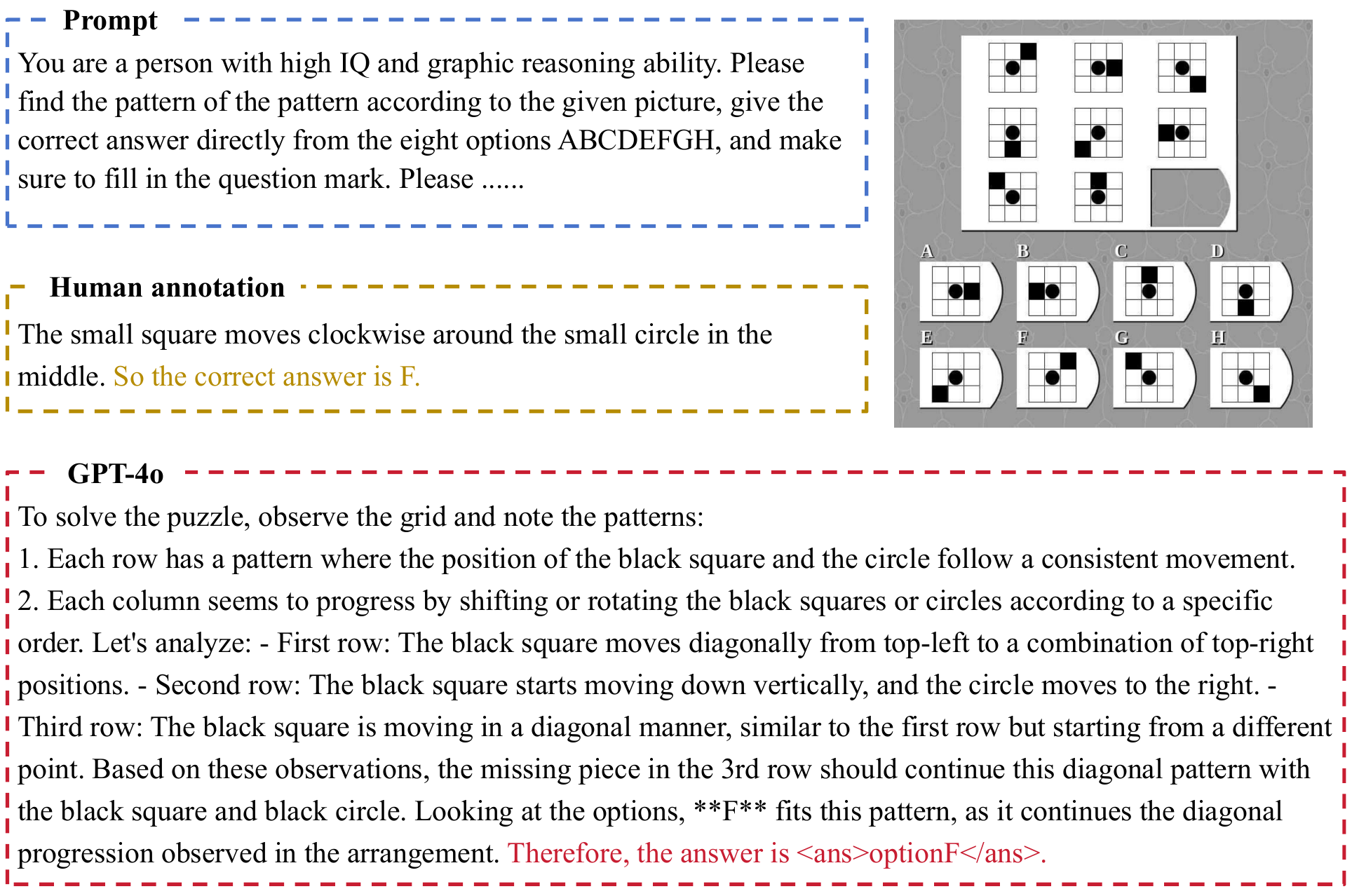}
  \caption{Example of the Mensa v2 task in the Mensa dimension.}
  \label{fig:example}
\end{figure*}

\begin{figure*}[htbp]
  \centering
  \includegraphics[width=\textwidth]{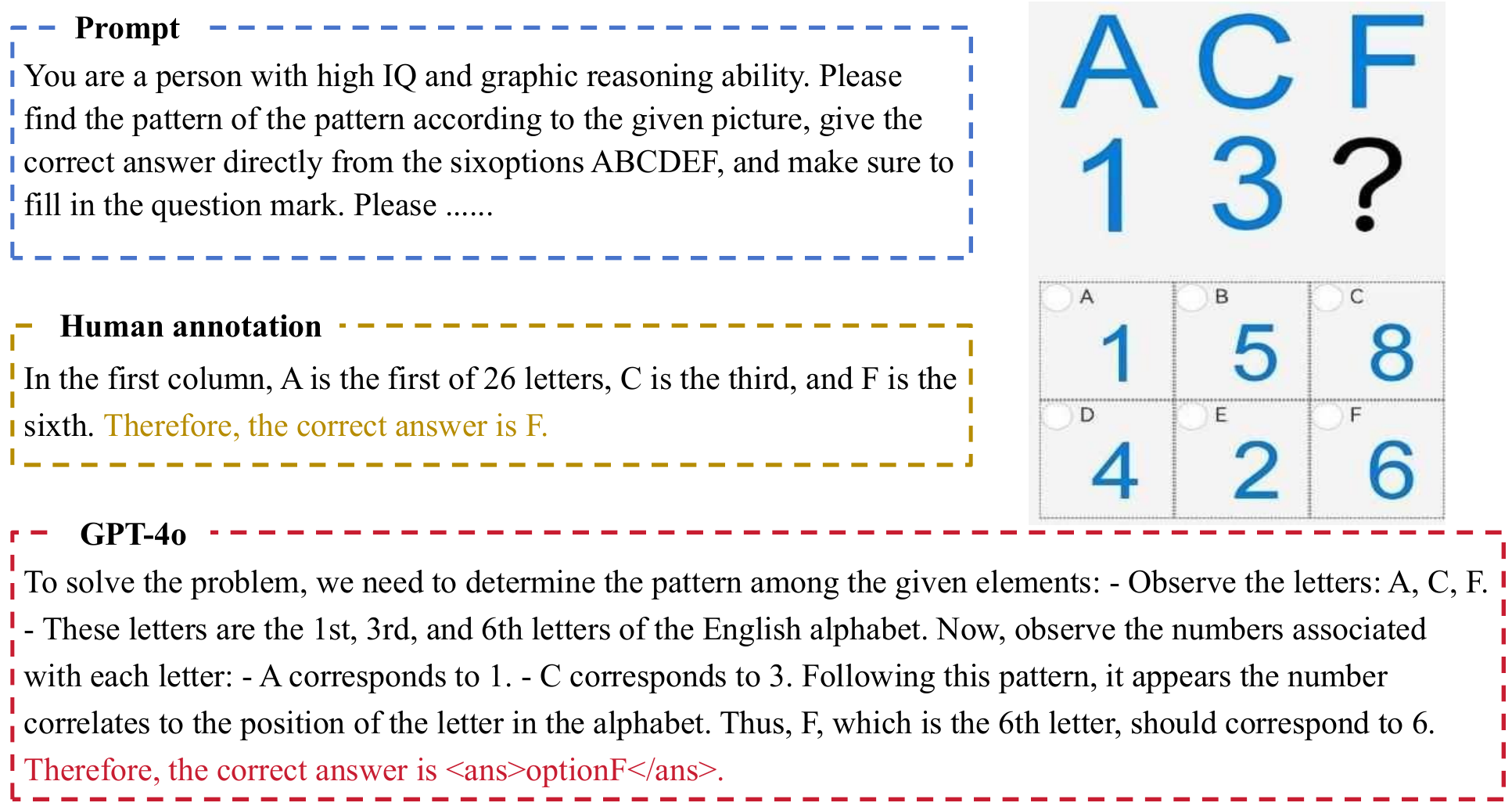}
  \caption{Example of the Raven v1 task in the Raven dimension.}
  \label{fig:raven40}
\end{figure*}

\begin{figure*}[htbp]
  \centering
  \includegraphics[width=\textwidth]{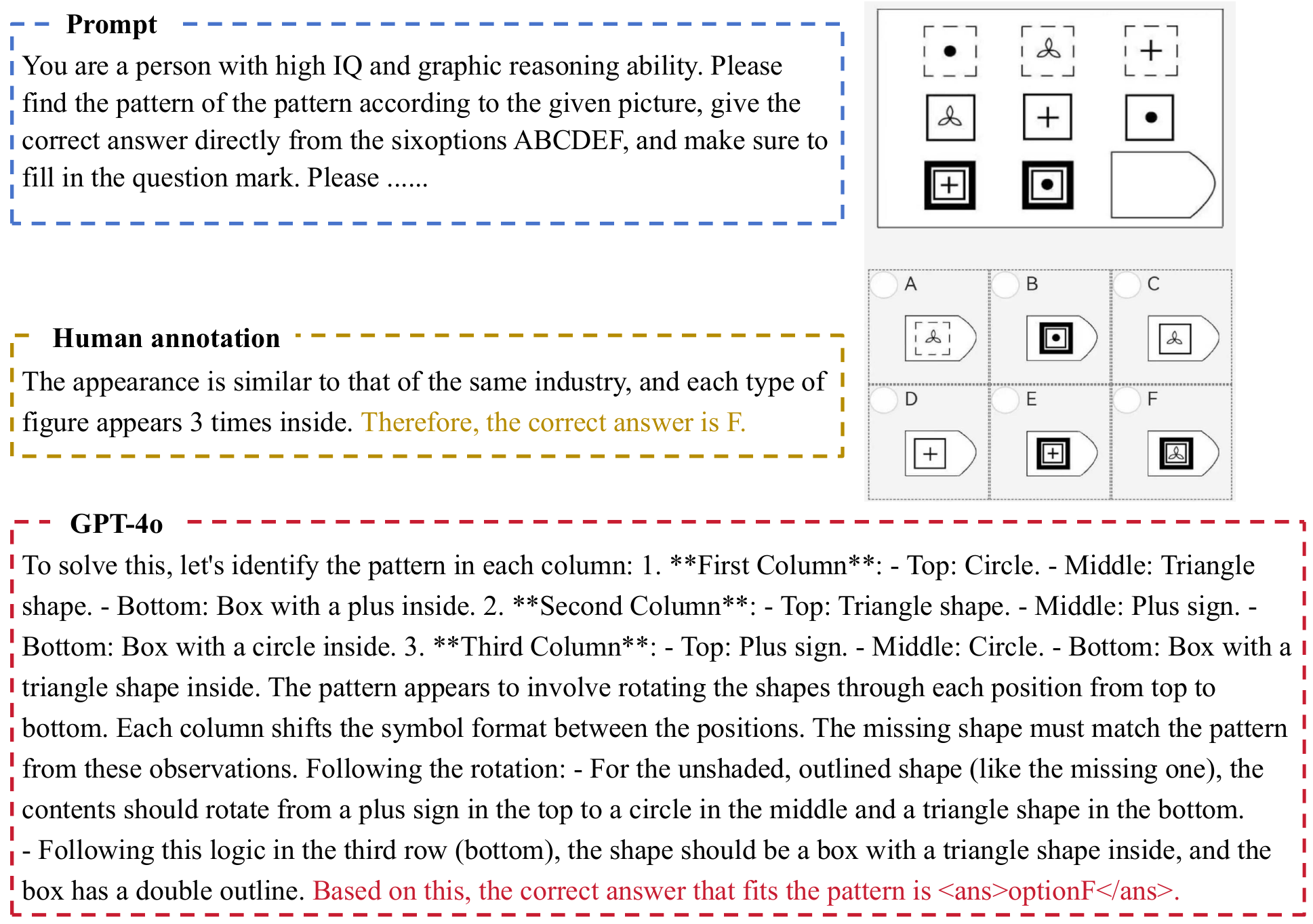}
  \caption{Example of the Raven v2 task in the Raven dimension.}
  \label{fig:raven60}
\end{figure*}

\end{document}